\documentclass[IJPHM, 2014, 00]{PHMSociety}

\usepackage{graphicx}
\usepackage{amsmath}
\usepackage{amssymb}
\usepackage{multirow}
\usepackage{adjustbox,lipsum, bm}
\usepackage{float}
\usepackage{cleveref}

\usepackage[usenames,dvipsnames]{color}

\begin{document}

\title{Constraint-Guided Learning of Data-driven Health Indicator Models: An Application on the Pronostia Bearing Dataset}

\author{%
Yonas Tefera\authorNumber{1, 2, 3, 4}, Quinten Van Baelen\authorNumber{1, 2, 3}, Maarten Meire\authorNumber{1, 2, 3}, Stijn Luca\authorNumber{5} and Peter Karsmakers\authorNumber{1, 2, 3}
}

\address{
    \affiliation{{1}}{KU Leuven, Dept. of Computer Science, ADVISE-DTAI, Kleinhoefstraat 4, B-2440 Geel, Belgium}
    \affiliation{{2}}{Leuven.AI - KU Leuven institute for AI}
    \affiliation{{3}}{Flanders Make @ KU Leuven}{ 
    {\email{\{yonas.tefera, quinten.vanbaelen, maarten.meire, peter.karsmakers\}@kuleuven.be}}
		} 
    \tabularnewline 
    \affiliation{{4}}{Addis Ababa University, School of Electrical and Computer Engineering, Addis Ababa, Ethiopia}{ 
    {\email{yonas.yehualaeshet@aau.edu.et}}
		} 
    \tabularnewline 
    \affiliation{5}{Ghent University, Department of Data Analysis and Mathematical Modelling, Coupure Links 653, 9000 Gent, Belgium}{ 
    {\email{stijn.luca@ugent.be}}
		} 
}

\maketitle
\pagestyle{fancy}
\thispagestyle{plain}

\phmLicenseFootnote{Yonas Tefera}

\begin{abstract}%
This paper presents a constraint-guided deep learning framework to develop physically consistent health indicators in bearing prognostics and health management. Conventional data-driven approaches often lack physical plausibility, while physics-based models are limited by incomplete knowledge of complex systems. To address this, we integrate domain knowledge into deep learning models via constraints, ensuring monotonicity, bounding output ranges between 1 and 0 (representing healthy to failed states respectively), and maintaining consistency between signal energy trends and health indicator estimates. This eliminates the need for complex loss term balancing to incorporate domain knowledge. We implement a constraint-guided gradient descent optimization within an autoencoder architecture, creating a constrained autoencoder. However, the framework is flexible and can be applied to other architectures as well. Using time-frequency representations of accelerometer signals from the Pronostia dataset, the constrained model generates more accurate and reliable representations of bearing health compared to conventional methods. It produces smoother degradation profiles that align with the expected physical behavior. Model performance is assessed using three metrics: trendability, robustness, and consistency. When compared to a conventional baseline model, the constrained model shows significant improvement in all three metrics. Another baseline incorporated the monotonicity behavior directly into the loss function using a soft-ranking approach. While this approach outperforms the constrained model in trendability, due to its explicit monotonicity enforcement, the constrained model performed better in robustness and consistency, providing stable and interpretable health indicator estimates over time. The results of the ablation study confirm that the monotonicity constraint enhances trendability, the boundary constraint ensures consistency, and the energy-health indicator consistency constraint improves robustness. These findings demonstrate the effectiveness of constraint-guided deep learning in producing reliable and physically meaningful health indicators for bearing prognostics and health management, offering a promising direction for future prognostic applications.
\end{abstract}

\section{Introduction}
\label{sec:intro} 
Prognostics and Health Management (\text{PHM}) is a domain that focuses on the management of the health of engineering systems and their critical components. It involves monitoring an asset's condition through sensors, collecting data to assess the health status, and predicting future health by analyzing these data. This insight is then used to enhance the asset's overall reliability and availability. \text{PHM} primarily addresses three key aspects \cite{Lei2018}: the construction of Health Indicator (\text{HI}), the prediction of the Remaining Useful Life (\text{RUL}) and the management of asset health. The construction of \text{HI}s is a crucial step in assessing and predicting the condition of assets \cite{biggio2020prognostics}. The goal is to extract effective features from sensor signals to establish stable metrics or indicators that reflect the asset's degradation over time. Preferably, these indicators should have a fixed range, such as 1 to 0, where 1 represents a healthy condition and 0 indicates a faulty one. This allows for the definition of a threshold that can trigger an alarm if the \text{HI} falls below it, providing an early warning before the system escalates into critical failure.

Bearings, as vital components of machinery in various industrial applications, are subjected to high stress and wear \cite{cubillo2016review}. As a result, an effective \text{PHM} for bearings is vital in preventing unexpected failures, improving operational efficiency, and prolonging the service life of the machines. Consequently, it is crucial to develop efficient techniques for monitoring bearing health. Various methods are employed in the industry to monitor bearing health and detect early signs of failure \cite{brkovic2017}. There are two main aspects that are important in these techniques \cite{Zhou2022}: (i) the detection of incipient bearing failures to avert unexpected fatal damage, and (ii) the tracking and characterization of the health condition's evolution over time. 

The modeling approaches used in the construction of \text{HI}s for bearings are broadly categorized into two types: physics-based methods and data-driven methods \cite{Li2024}. Physics-based methods \cite{xu2023physics} involve analyzing and modeling the physical failure mechanisms of bearings using fundamental principles. When these mechanisms or relevant domain knowledge are well understood and the parameters of these physical models can be accurately estimated from measurement data, physics-based methods can provide precise, generalizable and interpretable \text{HI} values. However, in most cases, particularly for complex engineering systems, the physical failure mechanisms and domain knowledge may be incomplete or scarcely available. In this situation, the established physical models must make simplifications or assumptions, which leads to a poor representative models \cite{lei2016model}. Thus, as an alternative, data-driven methods can be applied \cite{jieyang2023systematic}, which are particularly promising for tackling complex processes that are not entirely understood or when physics-based methods are too computationally demanding. Data-driven methods have relatively high data collection requirements and do not require extensive consideration of the physical meaning of the data \cite{Lu2024}. Although data-driven models can achieve high accuracy in fitting observed data, they often lack physical realism or plausibility when interpolating or extrapolating beyond the available labeled data, leading to poor generalization. It is clear that both physics-based and data-driven methods have their unique strengths and weaknesses. As a result, a sensible strategy is to combine these two types of methods to take advantage of their respective merits through a hybrid approach \cite{Liao2014}. Unlike the typical combination found in hybrid methods, embedding physical knowledge into data-driven models holds promise to guide models towards generating physically consistent results \cite{VonRueden2023}. Accordingly, there is a need to train data-driven machine learning (\text{ML}) models by incorporating physical or domain knowledge \cite{Karniadakis2021, Meng2022} to improve the interpretability of the model. This domain knowledge often comes in the form of physical models, constraints, dependencies, and known valid ranges of features \cite{Muralidhar2018}. There are various methods to incorporate specific physical or domain information into \text{ML} models, enabling the development of robust physics-informed \text{ML} systems. These methods include physics-informed data augmentation \cite{Xiong2023}, architecture design \cite{Nascimento2019, Chen2021}, residual modeling \cite{Willard2022}, and modifying the loss function to include a physics-inspired regularization term \cite{Wang2020, Raissi2019}. In the latter methods, the optimization procedure is adapted such that during learning models that are not consistent with the domain knowledge are penalized to guide learning towards a model that makes a trade-off between explaining the measurements and being consistent with domain knowledge. Adding domain knowledge to learning can be considered as a regularization method which can reduce the need for labeled data, shrink the search space during model optimization, and enhance the model's generalizability to unseen scenarios \cite{Li2024}. Using this strategy, the domain knowledge described by some formal representation, such as an inequality constraint, is transferred to the parameter values of a (black box) data-driven model. In our opinion, this strategy is one of the most convenient ways to embed knowledge in data-driven models; therefore, it is selected for designing \text{HI} models in this work.

In prior works, physics-inspired constraints were considered by adding an additional regularization term to the loss function. Such a term requires a proper trade-off hyperparameter, which requires careful balancing with the other terms in the objective function. Especially when more constraints are considered, which all require their own hyperparameter, such balancing can become cumbersome. In this work, a unified framework proposed by \citeA{VanBaelen2022} used to include domain knowledge in a data-driven approach to estimate the \text{HI} for bearings. This framework does not require a tuning of the hyperparameters of the different regularization terms and can be easily extended with the inclusion of additional domain-knowledge inspired constraints. More specifically, in this work the following constraints are considered: monotonicity, bounded ranges from fully healthy to failure states, and a characteristic degradation pattern informed by energy trends. This approach allows \text{HI} estimates to remain within defined ranges and adhere to the actual observed degradation patterns of a bearing, without the need to rely on a predefined shape of the degradation function. The proposed method will be experimentally tested and compared with two baseline \text{HI} model learning methods using the Pronostia data set. 

The remainder of this paper is organized as follows. In Section~\ref{sec:Methodology}, we present the proposed novel methodology that discusses the training approach, the proposed constraints, and the baseline methods used. Section~\ref{sec:Experimental_Setup} discusses the experimental setup starting with the general \text{HI} construction procedure, description of the data set, the pre-processing procedure, the model architecture used, the data set partitioning procedure and the evaluation metrics. In Section~\ref{sec:Experimental_Results}, we describe the implementation of our proposed approach and discuss the results. In addition, a comparison of the results with a conventional baseline analysis is implemented. An ablation study is presented in Section~\ref{sec:ablation} to discuss performance changes by slightly modifying the proposed model. Finally, in Section~\ref{sec:conclusion}, we conclude by summarizing the results of our experiments.

\section{Methodology}
\label{sec:Methodology}
The proposed method for constructing a bearing \text{HI} model leverages domain knowledge to guide the training process, resulting in a robust and accurate model. Incorporating domain knowledge during training is expected to reduce the reliance on labeled data. Unlike other approaches in the literature, domain knowledge in this work is not integrated by adding a regularization term to the loss function. Instead, the method employs the Constraint Guided Gradient Descent (\text{CGGD}) framework proposed in \cite{VanBaelen2022}, which facilitates solving constrained non-linear optimization problems. Specifically, this framework is utilized to train a Deep Learning (\text{DL}) model by minimizing a loss function while ensuring that the model satisfies certain constraints. These constraints represent domain knowledge. In this way, domain knowledge can be embedded into the resulting \text{DL} model. First, the application of the \text{CGGD} framework to develop \text{DL}-based \text{HI} models is described. Next, a detailed explanation of the domain-knowledge inspired constraints are provided along with their implementation within the \text{CGGD} framework. Finally, the baseline methods used for comparison with the proposed approach are explained.

\subsection{Constraint Guided Learning of a \text{DL} based \text{HI} Model}
\label{subsec:CGGD}
\text{DL} architectures commonly used in \text{PHM} \cite{Rezaeianjouybari2020} include Convolutional Neural Networks (\text{CNN}), Autoencoders (\text{AE}), Deep Belief Networks (\text{DBN}), Recurrent Neural Networks (\text{RNN}) and Generative Adversarial Networks (\text{GAN}). In systems where obtaining representative labeled data is challenging, the use of unsupervised learning methods, such as \text{AE} architectures, are particularly beneficial \cite{HoffmannSouza2023}. Although the proposed method is agnostic to the model architecture and learning procedure, in this work a Convolutional Autoencoder (\text{CAE}) model architecture learned by using a reconstruction loss is adopted as a starting point. As will be explained later in Section~\ref{sec:Experimental_Setup}, acceleration signals are transformed into a time-frequency representation denoted as $X \in \mathbb{R}^{D \times T}$ to be used as an input to the model.

In its original formulation, the \text{CAE} model is trained to construct a compact latent representation $z \in {\mathbb{R}}^{D'}$ of input signals $X$ with ${D'} \ll D$ using an encoder function $\mathcal{E}$. This process occurs without significant loss of information. Given $z$, a decoder $\mathcal{D}$ reconstructs the input to $\hat{X}$, aiming to closely resemble the original signal $X$. During learning, the model parameters are determined using the following objective \cite{vincent2010stacked}:
\begin{equation}\label{eq:CAE_Optm}
    \min_{\bm{\theta} _\mathcal{E},\bm{\theta} _\mathcal{D}} \mathcal{L}_{\text{reconn}}(\bm{X},\bm{\theta}_{\mathcal{E}},\bm{\theta}_{\mathcal{D}}),
\end{equation}
where $\bm{X}$ denotes the set of all input samples, \\
$\mathcal{L}_{\text{reconn}}(\bm{X},\bm{\theta}_{\mathcal{E}},\bm{\theta}_{\mathcal{D}})=\sum_{X \in \bm{X}} \Vert X - \mathcal{D}(\mathcal{E}(X)) \Vert _{2} ^{2}$ denotes the reconstruction loss, and $\bm{\theta} _\mathcal{E}$ and $\bm{\theta} _\mathcal{D}$ are the parameters of the encoder $\mathcal{E}$ and decoder $\mathcal{D}$ models, respectively.  The model is trained using data segments from a healthy bearing in operation, resulting in a minimal reconstruction error for this healthy state. In contrast, in scenarios where the bearing is faulty, the reconstruction process becomes less accurate, leading to a larger reconstruction error. This increased error serves as an indicator of bearing health degradation. Once the \text{CAE} parameters are learned, the \text{HI} estimate for an input $X$ is calculated using $f^{\text{CAE}}_{\text{HI}}(X)=-\|X-\mathcal{D}(\mathcal{E}(X))\|_2$ which computes the reconstruction error and will have a decreasing trend over time.

Using the \text{CGGD} framework, the optimization problem provided in Eq.~(\ref{eq:CAE_Optm}) can be turned into a constrained optimization task as:
\begin{equation}\label{eq:Const_Optm}
\begin{aligned}
    & \min_{\bm{\theta} _\mathcal{E},\bm{\theta} _\mathcal{D}} \quad \mathcal{L}_{\text{reconn}}(\bm{X},\bm{\theta} _{\mathcal{E}},\bm{\theta}_{\mathcal{D}}) \\
    & \hspace{2em}\text{s.t.} \quad C_i ({\bm{X}, \bm{\theta} _\mathcal{E},\bm{\theta} _\mathcal{D}}), \quad \text{for } i = 1, \ldots, M.
\end{aligned}
\end{equation}
where $C_i : \mathbb{R}^n \rightarrow \mathbb{R}$ is the $i$-th constraint, and $M$ is used to denote the number of constraints incorporated in the model. To apply \text{CGGD}, a direction $\operatorname{dir}_{\textbf{C}}$ needs to be defined for each constraint individually. This direction, when used to update the model, will result in a local improvement with respect to the constraint considered. For example, the constraint $C:\mathbb{R}^2\to\mathbb{R}:(x_1,x_2)\mapsto x_1-x_2$ with $C(x_1,x_2) \leq 0$ can be given a possible direction by $\operatorname{dir}_{\textbf{C}}= (\frac{\sqrt{2}}{2},-\frac{\sqrt{2}}{2})$. Consider, for example, an intermediate solution $(x_1,x_2)=(3,1)$ which does not satisfy the constraint. Observe that when the gradient of a gradient descent based optimizer is replaced by the direction $\operatorname{dir}_{\textbf{C}}$, $(x_1,x_2)=(3,1)-\eta(\frac{\sqrt{2}}{2},-\frac{\sqrt{2}}{2})$, that the updated solution, for a large enough $\eta$, will satisfy the constraint as $x_1$ can be decreased and $x_2$ can be increased until they at least have an equal value. 

To provide greater flexibility, this work employs a \text{CGGD} based \text{CAE} (\text{CCAE}), where the \text{HI} value is not derived from the reconstruction error, as in conventional \text{CAE}s, but is instead defined as a learnable function of the encoding $\mathcal{E}(X)$. This will be denoted as $f^{\text{CCAE}}_{\text{HI}}(\mathcal{E}(X))$ with the learnable parameters $\bm{\theta}_{\text{HI}}$.

\subsection{Constraints}
\label{subsec:constraints}
In this section, the constraints applied in this work are described to ensure that the predicted bearing \text{HI} values accurately reflect their degradation over time. These constraints are designed to represent domain knowledge to ultimately improve the accuracy and robustness of the model even in situations where annotated data is scarce. 

\subsubsection{Monotonic Degradation Constraint}
When a bearing is put into operation, it undergoes inevitable wear, progressively worsening over time. This degradation should be reflected in the predicted \text{HI}, which is expected to decrease monotonically over time. To enforce this constraint, the \text{HI} estimates should be penalized if they deviate from the expected monotonic trend based on the time location of the corresponding input data samples. For inputs $X_t$ measured at time step $t$ and the corresponding health state predictions $ f^{\text{CCAE}}_{\text{HI}}(\mathcal{E}(X_t))$, the monotonic degradation constraint can be expressed as follows:\\
For $t_1 < t_2$, hence $X_{t_1}$ a measurement taken before $X_{t_2}$ then it must hold that $ f^{\text{CCAE}}_{\text{HI}}(\mathcal{E}(X_{t_1})) > f^{\text{CCAE}}_{\text{HI}}(\mathcal{E}(X_{t_2}))$. To enforce such a constraint with \text{CGGD}, a direction function needs to be defined, which is done by comparing the ranks of $ f^{\text{CCAE}}_{\text{HI}}(\mathcal{E}(X_t))$ (from high to low value) estimates with their corresponding ranks in time (from early to later). This is calculated as follows:
\begin{multline}
\label{eq:Mono_Const}
\operatorname{dir}_{\text{mono}} (X_t,\bm{X},\bm{t},\bm{\theta}_\mathcal{E},\bm{\theta}_{\text{HI}}) = \\
\text{rank}_{\text{desc}} (X_t, \bm{X}, \bm{\theta}_\mathcal{E},\bm{\theta}_{\text{HI}}) - \text{rank}_{\text{asc}} (\text{time}(X_t), \bm{t}),
\end{multline}
where $\bm{t}$ is a set of timestamps aligned with the samples in set $\bm{X}$, $X_t$ is an element from set $\bm{X}$, $\text{time}(X_t)$ a function that extract the time at which the sample $X_t$ was measured, $\text{rank}_{\text{desc}}(\cdot)$ is a function that outputs the rank of $f^{\text{CCAE}}_{\text{HI}}(\mathcal{E}(X_t))$ when compared to all other \text{HI} estimates of the elements in $\bm{X}$ when ranked in descending order, $\text{rank}_{\text{asc}} (\cdot)$ is a function that outputs the rank of $t$ among all values in $\bm{t}$ sorted in ascending order.

As time progresses, it is expected that the corresponding \text{HI} estimates decrease. A positive value for $\operatorname{dir}_{\text{mono}}$ implies that the \text{HI} estimate is ranked too high, suggesting that a decrease in the \text{HI} estimate is necessary. In contrast, a negative value implies a lower rank, indicating that the \text{HI} estimate should be increased. A value of zero means that the \text{HI} estimate matches its expected rank. A significant deviation in the ranking of a sample from the desired position will have a greater impact on the direction than a smaller deviation. By incorporating this monotonic degradation constraint into the \text{HI} estimation process, the model ensures that the predicted \text{HI} values decrease monotonically over time, aligning with the expected degradation pattern of the bearing.

\subsubsection{Energy-\text{HI} Consistency Constraint}
Building on the monotonic degradation constraint, we expect that while the \text{HI} values should decrease over time, the difference between the \text{HI} values of two consecutive samples should not vary significantly unless a substantial change in signal energy occurs. To enforce this concept, a constraint is introduced that ensures that if two samples have similar energy levels, their \text{HI} estimates should also be close in value. 

We define this relationship mathematically as follows:
\begin{equation}
\label{eq:HI_Vs_Energy}
     f^{\text{CCAE}}_{\text{HI}}(\mathcal{E}(X_{t_0})) - \alpha \Delta \le  f^{\text{CCAE}}_{\text{HI}}(\mathcal{E}(X_{t_i})) <  f^{\text{CCAE}}_{\text{HI}}(\mathcal{E}(X_{t_0})),
\end{equation}
where $\Delta = \max \left(\kappa, \vert E(X_{t_i}) - E(X_{t_0}) \vert \right)$ represents the normalized total energy\footnote{The energy $E(\cdot)$ is calculated by summing all squared elements in $X$.} difference between the two samples at times $t_0$ and $t_i$ where $t_i > t_0$, $\alpha > 0$ is a hyperparameter that controls the sensitivity of this constraint and $0 < \kappa < 1$. $\kappa$ is a parameter that introduces flexibility in the \text{HI} predictions between samples that exhibit similar energy values. Specifically, when the energy difference between the samples is minimal, $\kappa$ allows a margin of flexibility in the \text{HI} predictions.

Based on the predicted \text{HI} of the model, the subsequent update direction of the energy-\text{HI} consistency constraint is determined as follows:
\begin{multline}    
\hspace{-10pt}\operatorname{dir}_{\text{ene}}( X_{t}, X_{t_0}, \bm{\theta}_\mathcal{E},\bm{\theta}_{\text{HI}}) = \\
\begin{cases}
    1, \quad f^{\text{CCAE}}_{\text{HI}}(\mathcal{E}(X_{t})) >  f^{\text{CCAE}}_{\text{HI}}(\mathcal{E}(X_{t_0})), \\
    0, \quad -\alpha \Delta \le  f^{\text{CCAE}}_{\text{HI}}(\mathcal{E}(X_{t})) -  f^{\text{CCAE}}_{\text{HI}}(\mathcal{E}(X_{t_0})) < 0, \\
   -1, \quad \text{otherwise}.
\end{cases}
\end{multline}
By penalizing discrepancies between the energy difference and the \text{HI} difference, we encourage the model to maintain a consistent relationship between these two variables. This approach helps prevent large fluctuations in the \text{HI} values and ensures that the model accurately reflects the gradual degradation process of the bearing. 

\subsubsection{\text{HI} Boundary Constraint}
When predicting the \text{HI} of a bearing, it is convenient that the values remain within a normalized range: a fully healthy state is represented by a value of $ub = 1$, while a failure state corresponds to a value of $lb = 0$. To enforce this condition, boundary constraints are enforced during the training process to ensure that all \text{HI} predictions fall within this defined range. 

In addition to the broader bound ranges, stricter bound ranges are also applied during certain phases of the bearing's life cycle. In the initial $a \%$ (e.g. $a = 10$) of its operation, the \text{HI} value is expected to be at least $b_a < ub$ (e.g. $b_a = 0.9$), ensuring that the new bearings operate in near-optimal health. In contrast, in the final $b\%$ (e.g. $b = 5$) of its operation, the \text{HI} is expected not to exceed $b_b > lb$ (e.g. $b_b = 0.05$), indicating that the bearing is close to failure due to significant degradation.

Based on the predicted \text{HI}, the subsequent upper and lower boundary constraints direction is determined as follows:
\[
\operatorname{dir}_{\text{upper}}(X_{t}, \bm{\theta}_\mathcal{E},\bm{\theta}_{\text{HI}}) = 
\begin{cases}
    1, & \text{if } f^{\text{CCAE}}_{\text{HI}}(\mathcal{E}(X_{t})) > ub, \\
    0, & \text{otherwise},
\end{cases}
\]
\[
\operatorname{dir}_{\text{lower}}(X_{t}, \bm{\theta}_\mathcal{E},\bm{\theta}_{\text{HI}}) = 
\begin{cases}
    -1, & \text{if } f^{\text{CCAE}}_{\text{HI}}(\mathcal{E}(X_{t})) < lb,\\
    0, & \text{otherwise},
\end{cases}
\]
where $lb$ and $ub$ are the lower and upper bounds, respectively, within which the \text{HI} predictions should lie.

When $\operatorname{dir}_{\text{upper}}$ is 1, it means that the \text{HI} estimate is greater than $ub$, which requires a reduction in subsequent updates. When $\operatorname{dir}_{\text{lower}}$ is -1, the \text{HI} estimate is below $lb$ and should be increased. A zero in both $\operatorname{dir}_{\text{upper}}$ and $\operatorname{dir}_{\text{lower}}$ indicates that the \text{HI} estimate is within the specified bounds and that no adjustments are necessary. By implementing these boundary constraints, it is ensured that the \text{HI} predictions accurately reflect the bearing condition, from optimal functionality to failure.

\subsection{Implementing the Constraints in the \text{CGGD} Framework}
This section provides a comprehensive explanation on how the \text{CGGD} framework can be used to train the \text{HI} estimator model, denoted as $ f^{\text{CCAE}}_{\text{HI}}(\mathcal{E}(\cdot))$. As is indirectly indicated in Eq,~(\ref{eq:Const_Optm}), a multi-head network model is used. One head decodes the encoded input to calculate the reconstruction loss (to calculate $\mathcal{L}_{\text{reconn}}$) and another head is used for the \text{HI} prediction ($ f^{\text{CCAE}}_{\text{HI}}$) based on the encoded input. Both will use the same encoder, and the input of each head is thus given by $\mathcal{E}(X)$. 

At the core of the \text{CGGD} optimization procedure, the update of the model parameters is defined in Eq.~(\ref{eq:CGGDUsageSingleColumn}).
\begin{figure*}[htbp]
\hrule
    \begin{align}
        \label{eq:CGGDUsageSingleColumn}
        \theta_{j+1} := \theta_j - \eta \left( \frac{\partial \mathcal{L}_{\text{reconn}}\left(X_t, \bm{\theta}_{\mathcal{E}}, \bm{\theta}_{\mathcal{D}} \right)}{\partial \theta_j} + \right. & \left. \max\left\{ \left\|\nabla_{\mathcal{E}} \mathcal{L}_{\text{reconn}} \left(X_t, \bm{\theta}_{\mathcal{E}}, \bm{\theta}_{\mathcal{D}} \right)\right\|,\epsilon\right\} \frac{\partial f^{\text{CCAE}}_{\text{HI}}\left(\mathcal{E}\left(X_t\right)\right)}{\partial \theta_j} \right. \\ \nonumber
        & \Bigl[R_{\text{mono}} \operatorname{dir}_{\text{mono}} \left(X_t, \bm{X}, \bm{t}, \bm{\theta}_{\mathcal{E}}, \bm{\theta}_{\text{HI}}\right) F_{\text{MH}} \left(X_t,\operatorname{dir}_{\text{mono}}(X_t, \bm{X}, \bm{t}, \bm{\theta}_{\mathcal{E}}, \bm{\theta}_{\text{HI}})\right) \\ \nonumber 
        & \quad + R_{\text{ene}} \operatorname{dir}_{\text{ene}} \left(X_t, X_{t_0}, \bm{\theta}_{\mathcal{E}}, \bm{\theta}_{\text{HI}}\right) F_{\text{MH}} \left(X_t,\operatorname{dir}_{\text{ene}}(X_t, X_{t_0}, \bm{\theta}_{\mathcal{E}}, \bm{\theta}_{\text{HI}})\right) \\ \nonumber 
        & \quad + R_{\text{upper}} \operatorname{dir}_{\text{upper}} \left(X_t, \bm{\theta}_{\mathcal{E}}, \bm{\theta}_{\text{HI}}\right) F_{\text{MH}} \left(X_t,\operatorname{dir}_{\text{upper}}(X_t, \bm{\theta}_{\mathcal{E}}, \bm{\theta}_{\text{HI}})\right) \\ \nonumber 
        & \quad + R_{\text{lower}} \operatorname{dir}_{\text{lower}} \left(X_t, \bm{\theta}_{\mathcal{E}}, \bm{\theta}_{\text{HI}}\right) F_{\text{MH}} \left(X_t,\operatorname{dir}_{\text{lower}}(X_t, \bm{\theta}_{\mathcal{E}}, \bm{\theta}_{\text{HI}})\right) \Bigr] \Bigr)
    \end{align}
\hrule
\end{figure*}
Here $\theta_j$ is the value of a learnable weight on iteration $j$ and $\theta_{j+1}$ the value of the same learnable weight in the next iteration, $R_{\text{mono}}$, $R_{\text{ene}}$, $R_{\text{upper}}$, and $R_{\text{lower}}$ are the rescale factors, which control the relative weight of the different constraints compared to the loss function, for the monotonic degradation constraint, the energy-\text{HI} consistency constraint, the upper bound on the \text{HI}, and the lower bound on the \text{HI}, respectively, $\eta$ is the learning rate, $F_{\text{MH}}$ is a function that balances the gradients of the different heads appropriately, $\nabla_\mathcal{E}$ is the gradient with respect to the latent space determined by the encoder $\mathcal{E}$, and $0<\epsilon<1$ to ensure there is a minimum step size in case the gradients from the reconstruction loss to the encodings are very small. A small adaptation to standard \text{CGGD} can be observed as the variable $F_{\text{MH}}$ is included. This is required as $\bm{\theta}_{\text{HI}}$ are only trained using the constraints, while $\bm{\theta}_{\mathcal{D}}$ are only trained with a loss function. Therefore, the first shared space by both objectives is the encoding space. In this space, the constraints should be prioritized over the reconstruction loss function, as the final model should satisfy the constraints on the training data. The constraints and the loss function will determine an update vector for the encoding space. In order to make sure that the update vector linked to the constraints dominates the model updating both update vectors are set to the same norm. This is accomplished by first setting the update vector created by the constraints to have a unit norm, after which it is multiplied by the norm of the update vector that is calculated based on the loss function. To have a unit norm for the constraints-based update vector, first the gradient of the direction function ($\operatorname{dir}$) is multiplied by the automatic differentiation of the predicted \text{HI} ($ f^{\text{CCAE}}_{\text{HI}}(\mathcal{E}(X))$) for each encoding space dimension. Then, to let it have unit norm this update vector is multiplied by the rescale factor of the constraint and $F_{\text{MH}}$ which is defined as:
\begin{align*}
    F_{\text{MH}} & \left(X_t, \operatorname{dir}(X_t, \cdot)\right) := \\
    &\frac{\left\| \operatorname{dir}(X_t,\cdot) \right\|}{\left\| \nabla_{\mathcal{E}} \left[ \operatorname{dir}(X_t,\cdot)  f^{\text{CCAE}}_{\text{HI}}\left(\mathcal{E}\left(X_t\right)\right) \right]\right\|}.
\end{align*}

After being rescaled with $F_{\text{MH}}$, the constraint update vector is multiplied by $\|\nabla_{\mathcal{E}} \mathcal{L}_{\text{reconn}} \left(X_t, \bm{\theta}_{\mathcal{E}}, \bm{\theta}_{\mathcal{D}} \right)\|$, as can be seen in Eq.~(\ref{eq:CGGDUsageSingleColumn}), to let both the constraint and loss update vector have the same norm.

To train the neural network, an off-the-shelve Adam optimizer is used by applying Eq.~(\ref{eq:CGGDUsageSingleColumn}) iteratively for all learnable parameters. Observe that the partial derivatives of the loss function with respect to $\bm{\theta}_{\text{HI}}$ are 0 as they are independent from each other and the partial derivatives of the encodings with respect to $\bm{\theta}_{\mathcal{D}}$ are 0 as well as the weights of the decoder are independent from the encodings. 

All rescale factors are assumed to be strictly greater than 1, as they represent the relative weight of the constraints compared to the loss function by the definition of \text{CGGD}. Different values can be assigned to individual rescale factors to prioritize certain constraints over others. Generally, the constraint with the largest rescale factor takes precedence over all other constraints and the loss function.

In this work, the rescale factor for the monotonic degradation constraint is dynamically adjusted for each individual prediction based on its deviation from the ground truth, within a bounded interval defined by $[R_{\text{mono\_lw}}, R_{\text{mono\_up}}]$. This interval ensures higher rescale factors for larger deviations between the predicted and true ranks and comparatively lower factors for smaller deviations. For a monotonicity direction ($\operatorname{dir}_{\text{mono}}(X_t, \bm{X}, \bm{t}, \bm{\theta}_\mathcal{E}, \bm{\theta}_{\text{HI}})$) calculated for sample $X_t$, the monotonic rescale factor is calculated as:
\begin{align*}
    R_{\text{mono}_t} = & R_{\text{mono\_lw}} + \\
    &\frac{(R_{\text{mono\_up}} - R_{\text{mono\_lw}}) \operatorname{dir}_{\text{mono}}(X_t, \bm{X}, \bm{t}, \bm{\theta}_\mathcal{E}, \bm{\theta}_{\text{HI}})}{\text{batch\_size} - 1}.
\end{align*}

Here, $\text{batch\_size} - 1$ represents an upper bound on the absolute value of the values in $\operatorname{dir}_{\text{mono}}$. In particular, this upper bound can be attained when all samples are from the same run and the earliest sample is predicted as the last sample or the other way around.

\subsection{Comparison Baselines}
To evaluate the effectiveness of the proposed \text{CGGD}-based approach, it was compared to two baseline methods. The first baseline is a standard \text{CAE} method in which the encoder and decoder parameters are learned according to Eq.~(\ref{eq:CAE_Optm}). Hence, there is no additional head $f^{\text{CCAE}}_{\text{HI}}$ and there are no added constraints.

The second baseline incorporates the monotonic degradation property using a regularization term in the loss function of the model, unlike the proposed \text{CCAE} method, which applies it as a constraint during learning. However, if conventional ranking operations are used for the monotonic degradation loss function, they will create discrete, piecewise-constant outputs, like integer ranks, that are not differentiable. This poses challenges for gradient backpropagation in \text{DL} due to null or undefined derivatives. To overcome these issues, the method proposed by \citeA{Blondel2020} was used. 

The method casts the ranking as a projection onto the permutahedron (the convex hull of all permutation vectors). As a result, it creates projection operators that are differentiable, making them suitable for formal analysis. For a given input vector ($\bm{x}$), the soft rank $r_\varepsilon (\bm{x})$ is computed as:
\[
r_{\varepsilon} (\bm{x}) = Proj_{\mathcal{P}} (-\bm{x}/\varepsilon)
\]
where $\mathcal{P}$ is the permutahedron and $\varepsilon > 0$ is a regularization strength controlling approximation smoothness.

For the predicted \text{HI}s ($f_{\text{HI}}(\bm{X})$) of set $\bm{X}$, the soft-rank loss is calculated as:
\begin{equation}
    \mathcal{L}_{\text{soft-rank}} = \frac{1}{2} \Vert r_{\varepsilon} (\text{rank}_{\text{asc}}(\bm{t})) - r_{\varepsilon} (f_{\text{HI}}(\bm{X})) \Vert _{2} ^{2},
\end{equation}
where $\bm{t}$ is a set of timestamps aligned with the samples in set $\bm{X}$, $\text{rank}_{\text{asc}}(\bm{t})$ outputs the rank of  $\bm{t}$ sorted in ascending order which represents the true ranks and $r_{\varepsilon} (\cdot)$ provides the soft-ranks for the input vectors.

Then, the total loss function of the model is calculated as:
\begin{equation}
\label{eq:total_loss}
    \mathcal{L}_{\text{total}} = \mathcal{L}_{\text{reconn}} + \lambda \mathcal{L}_{\text{soft-rank}},
\end{equation}
where $\lambda$ is a trade-off hyperparameter that balances the reconstruction and the soft-rank term.

This second baseline, which incorporates monotonicity as a regularization term in the loss function using the soft-rank approach, is called the soft-rank loss function based \text{CAE} (\text{SR-CAE}) method.

\section{Experimental Setup: Bearing \text{HI} Estimation}
\label{sec:Experimental_Setup}
This section discusses the dataset, the preprocessing and partitioning approach, the model architecture, the setting of the hyperparameters and the relevant evaluation metrics used in this work.

\subsection{\text{HI} Construction Procedure}
\label{subsec:HI_Construction}
The framework for building a predictive model involves several steps, starting with collecting data from monitoring signals such as temperature and vibration to reflect the health of the equipment. This data is then pre-processed to enhance its quality for analysis. An appropriate model is then chosen based on the characteristics of the data and the prognostic needs, with a specific architecture designed to detect bearing degradation patterns. Finally, the model estimates the \text{HI}s of bearing health, providing insight into degradation over time.

\subsection{Pronostia Bearing Degradation Dataset}
This work uses the IEEE \text{PHM} 2012 Prognostic Challenge dataset, sourced from the Pronostia platform \cite{Nectoux2012}. The bearings on this platform are tested under various loads and rotational speeds, comprising $17$ run-to-failure datasets of rolling element bearings.

Table~\ref{Tab:Pronostia_desrc} summarizes the different conditions and the number of datasets available for training and testing purposes. To perform accelerated degradation tests within a few hours, a high-level radial force was applied that exceeded the maximum dynamic load of the bearings. During these tests, the rotating speed of each bearing was maintained at a stable level. Two accelerometers and a thermocouple were employed to capture vibration signals and bearing temperatures. Vibration signals were recorded using two accelerometers positioned along the vertical and horizontal axes, with a sampling frequency of $25.6 \text{kHz}$ and $2560$ samples (that is, $1/10$ s) collected every $10$ seconds.

Since the bearings were subjected to natural degradation, we expect that the degradation patterns will vary between samples. Furthermore, little is known about the specific nature and origin of degradation (for example, whether it involves balls, inner or outer races, or cages), necessitating the application of data-driven techniques \cite{Nectoux2012}. Failures in any component, ball, rings, or cage could occur simultaneously. The useful life of a bearing is considered to end when the amplitude of the vibration signal exceeds $20 g$. In this work, only accelerometer data from the provided data set were used as input to estimate the \text{HI}.
\begin{table}[htbp]
\renewcommand{\arraystretch}{1.25} 
\begin{center}
\adjustbox{max width=0.47\textwidth}{
\begin{tabular}{c||c|c|c|c} \hline \hline
Conditions & Load (N)  & Speed (rpm) & 
\begin{tabular}{@{}c@{}}Number of\\ Train\\ Bearings\end{tabular} & 
\begin{tabular}{@{}c@{}}Number of\\ Test\\ Bearings\end{tabular}\\\hline \hline
1 & 4000 & 1800 & 2 & 5\\ 
2 & 4200 & 1650 & 2 & 5\\ 
3 & 4500 & 1300 & 2 & 1\\\hline
\end{tabular}}
\caption{Summary of the Pronostia dataset.}
\label{Tab:Pronostia_desrc}
\end{center}
\end{table}

\subsection{Preprocessing}
Various vibration analysis techniques can be applied to preprocess the raw signals from an accelerometer \cite{Vishwakarma2017}. In this work a similar approach to \citeA{Meire2022} was used, where rich features were calculated that do not include prior knowledge about the bearings. Here, first the acceleration signals are transformed into \text{log-mel} spectrograms before using them as input to the machine learning model. The pronostia platform captured recordings of $0.1$ seconds or $2560$ samples every $10$ seconds. To extract the mel spectrograms, we processed the vertical and horizontal accelerometer axes using a window size of $0.1$ seconds and a hop size of $0.1$ seconds. We extract $128$ mel bands from each axis, apply logarithmic scaling, and then stack the resulting spectrograms. This creates a 2D input frame with dimensions $(128, 2)$ for each $0.1$-second window. 

\begin{figure}[htbp]
    \centering
    \includegraphics[scale=.6]{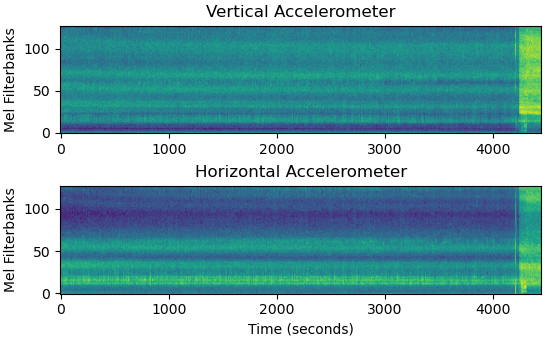}
    \caption{The two axes mel spectral features of a bearing.}
    \label{fig:MelFeatures}
\end{figure}

Figure~\ref{fig:MelFeatures} shows a sample of the extracted mel spectrograms from the first bearing that operates under condition 3. At the start of each bearing run, their is a transient behavior likely caused by the initial start-up of the setup. To eliminate this effect, we visually inspect and remove a small section of samples at the beginning of each bearing operation, as demonstrated in Figure \ref{fig:MelFeatures}.

Finally, we normalize the data from each operating condition to achieve zero mean and unit variance across the mel frequency bands, treating each individual axis separately. This normalization is based on the mean and standard deviation calculated from the training set. This normalization helps ensure consistency between different runs under the same conditions. 

By transforming raw vibration signals into \text{log-mel} spectrograms and applying the appropriate preprocessing steps, we obtain a consistent set of input features to train machine learning models for bearing \text{HI} estimation.

\subsection{Model Architecture}
\label{subsec:Modeling}
\begin{figure}[htbp]
    \centering
    \includegraphics[width=8.5cm, height=5.25cm]{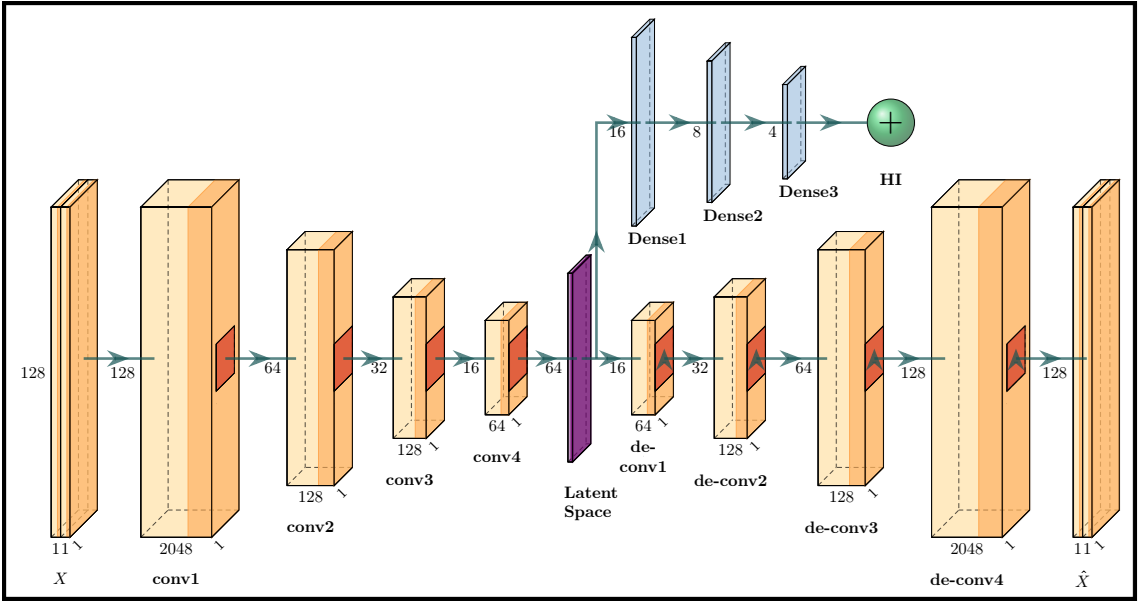}
    \caption{The \text{CCAE} Model Architecture.}
    \label{fig:arch}
\end{figure}

The standard \text{CAE} architecture used in this study, illustrated in Figure~\ref{fig:arch}, consists of an encoder with four convolutional layers containing 64, 32, 32, and 16 filters, respectively, followed by a fully connected layer with 16 neurons. The decoder consists of a fully connected layer and 4 deconvolutional layers with 16, 32, 32, and 64 filters, ending with a final deconvolutional layer with two filters, corresponding to the two-axis accelerometer data as the output layer. A batch normalization layer follows each convolutional layer, except the final one. All convolutional layers employ ReLu activation functions and the dense layers use linear activation functions. The convolutional filters used are 1D with a kernel size of 3 and move with a stride of 2. In \text{CCAE} a second head is added on top of the encoder output (upper head in Figure~\ref{fig:arch}) to estimate the \text{HI}. This second head is composed of three fully connected layers that have 16, 8, and 4 neurons, and the final layer providing a single \text{HI} estimate. The hyperparameters used in this model include the Adam optimizer with a learning rate of $1\times10^{-3}$, an early stopping criterion with a patience of 10 epochs, a batch size of 64. To ensure a standardized comparison of the methodologies, neither regularization nor dropout techniques are implemented in the architecture used.

The final architecture and hyperparameters are selected after conducting multiple experiments to fine-tune them. This is done through a systematic evaluation of various configurations based on reconstruction error and validation accuracy.

\subsection{Data Set Partitioning}
\begin{figure}[htbp]
    \centering
    \includegraphics[width=8.5cm, height=3.5cm]{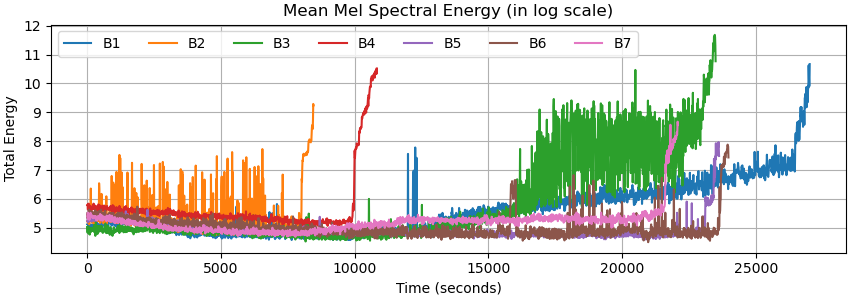}
    \caption{Mean Mel Energy Progression Over Time for Bearings in Operational Condition 1.}
    \label{fig:Spectral_Energy}
\end{figure}

In this work, two bearing runs are used to train the model, while the remaining runs are reserved for testing under each operational condition, as specified in the original \text{PHM} challenge \cite{Nectoux2012}. The training set for each condition is relatively small due to the limited number of runs available. Furthermore, fault patterns and bearing lifetimes can vary significantly, even under identical conditions, complicating the \text{HI} estimation. This variability is illustrated in Figure~\ref{fig:Spectral_Energy}, which shows the degradation processes of seven bearings that operate under a similar condition. Although the energy of these bearings is expected to follow a generally increasing trend, the progression towards failure varies considerably among them. Notably, certain bearings, such as B2 and B3, exhibit large energy fluctuations rather than a smooth transition.

From the \text{log-mel} spectral features, input batches are created in a structured manner. First, we determine how many samples from each run will comprise a batch, ensuring that each run contributes proportionally to the overall batch size. To enhance representativeness, each run is divided into three sections.
\begin{enumerate}
    \item Healthy State: The first section comprises the initial $10\%$ of the run, assumed to reflect a fully healthy state.
    \item Slight Degradation Phase: The second section includes samples from $10\%$ to $95\%$ of the run, representing a phase of slight degradation.
    \item Sharp Degradation Phase: The final section, expected to roughly begin after $95\%$  of the samples in a run, characterizes rapid degradation leading up to apparent failure. 
 \end{enumerate}
 
The batches are then created by randomly sampling from each section of the training runs used. Each batch is composed of $20\%$, $70\%$, and $10\%$ samples from the healthy state, the slight degradation phase, and the sharp degradation phase, respectively. This method ensures that batches include a balanced representation of data from all phases of degradation, enhancing the training process. To evaluate the performance of each \text{HI} estimation method, each experiment was repeated 10 times using different random seeds for batch generation and model initialization. For each scenario, data from the two selected training bearings were combined and shuffled. Of this combined dataset, $75\%$ is used for training, while the remaining $25\%$ is set aside for validation.

\subsection{\text{HI} Evaluation Metrics}
\label{subsec:HI_evaluation}
The effectiveness of the constructed \text{HI} estimates are assessed using three key metrics: trendability, robustness, and consistency \cite{Lei2018}.

\subsubsection{Trendability}
As operating time increases, components are expected to gradually degrade. Consequently, the degradation trend of an \text{HI} should correlate with the operating time. The trendability metric is used as a quantitative measure to evaluate how well \text{HI} reflects changes in the condition of a machine over time. To evaluate this correlation in nonlinear degradation trends, we use the Spearman coefficient as the trendability metric, defined as follows \cite{Carino2015}:
\begin{equation}
\label{eq:trendability}
    \text{Tre}(f_{\text{HI}}(\bm{X}), \bm{t}) = 1 - \frac{6\sum_{i=1}^N (\text{rank}(f_{\text{HI}}(X_{i})) - \text{rank}(t_i))^2}{N(N^2 - 1)},
\end{equation}
where \(N\) is the number of samples, \(\text{rank}(f_{\text{HI}}(X_{i}))\) and \(\text{rank}(t_i)\) are the ranks of the estimated \text{HI} values and the corresponding time value, respectively. The value of \(\text{Tre}(f_{\text{HI}}(\bm{X}), \bm{t})\) ranges from -1 to 1, approaching either end when there is a strong positive or negative correlation between \text{HI} and time. \(\text{Tre}(f_{\text{HI}}(\bm{X}), \bm{t}) = 1\) means the health indicator is perfectly increasing with time (positive trend). \(\text{Tre}(f_{\text{HI}}(\bm{X}), \bm{t}) = -1\) means the health indicator is perfectly decreasing with time (negative trend) and \(\text{Tre}(f_{\text{HI}}(\bm{X}), \bm{t}) = 0\) indicates that there is no trend. The aim of the constraint is to achieve a value of -1.

\subsubsection{Robustness}
A suitable \text{HI} should also be robust against inherent interference while maintaining a smooth degradation trend. This property is quantified by the robustness metric \cite{Zhang2016}, defined as:
\begin{equation}
\label{eq:robustness}
\begin{split}
    \text{Rob}&(f_{\text{HI}}(\bm{X}),\bm{t}) = \\
    &\frac{1}{N}\sum_{i=1}^{N}\operatorname{exp}\left(-\left\vert \frac{f_{\text{HI}}(X_i) - f^{s}_{\text{HI}}(t_i,\bm{X},\bm{t})}{f_{\text{HI}}(X_i)}\right\vert \right),
\end{split}
\end{equation}
where \(f^{s}_{\text{HI}}(t_i,\bm{X},\bm{t})\) is a smoothed \text{HI} value at \(t_i\). The smoothing is performed in this work using locally weighted regression (\text{LOESS}) \cite{Cleveland1988, Duong2018} where each smoothed value is determined using neighboring data points within a specified range.

\subsubsection{Consistency}
Consistency refers to the degree of correlation among multiple \text{HI}s. When examining different \text{HI} estimates from a single unit, it is expected that they will exhibit some level of correlation because they all reflect the same degradation process. This metric is especially useful in situations where multiple predictions are made, as it enables the assessment of the consistency between these predictions.\\
\citeA{mosallam2016data} proposed a consistency metric based on the pairwise symmetrical uncertainty, defined as:
\begin{equation}
\label{eq:consistency}
    \text{Con}(f_{\text{HI}}(\bm{X_1}), f_{\text{HI}}(\bm{X_2})) = \frac{2I(f_{\text{HI}}(\bm{X_1}), f_{\text{HI}}(\bm{X_2}))}{\text{H}(f_{\text{HI}}(\bm{X_1})) + \text{H}(f_{\text{HI}}(\bm{X_2}))},
\end{equation}
Where, $I(f_{\text{HI}}(\bm{X_1}),f_{\text{HI}}(\bm{X_2}))$ represents the mutual information and $\text{H}(f_{\text{HI}}(\bm{X_1}))$ and $\text{H}(f_{\text{HI}}(\bm{X_2}))$ denote the entropies of $f_{\text{HI}}(\bm{X_1})$ and $f_{\text{HI}}(\bm{X_2})$, respectively. The output value, $\text{Con}(f_{\text{HI}}(\bm{X_1}),f_{\text{HI}}(\bm{X_2}))$, is normalized to a range between 0 and 1. A higher value indicates a greater similarity between the two \text{HI}s, suggesting a stronger consistency. For multiple \text{HI} estimates, pairwise consistency values are computed and the final consistency metric is determined by taking the mean and standard deviation of these pairwise results.

\section{Evaluation of \text{HI} Estimation Methods: Experimental Results and Discussion}
\label{sec:Experimental_Results}
This section presents the experimental results, starting with a comparative analysis of the proposed \text{CCAE} method against two baselines: the standard \text{CAE} and the \text{SR-CAE} methods. This is followed by a detailed discussion of all methods based on the performance metrics described in Section~\ref{subsec:HI_evaluation}. When comparing results, if the models show the same mean performance metric, the one with a smaller standard deviation is considered better. For each bearing under the three operating conditions, the first two bearing data are used for training, while the remaining bearing data are set aside for testing.

\subsection{General Comparative Analysis}
The overall performance of each approach is visually assessed by analyzing the results of the bearings in Condition 3 across the three \text{HI} estimation methods. This analysis provides a general overview, providing insight into how effectively each technique estimates the health degradation of bearings. In all HI estimation plots presented in this work, the smoothing (red line) is done using \text{LOESS}.

\begin{figure*}[htbp]
    \centering
    \includegraphics[width=15.5cm, height=3.5cm]{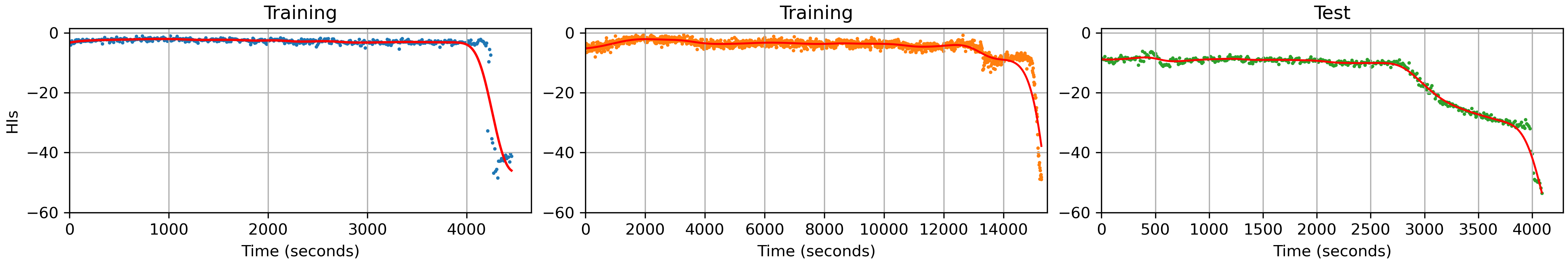}
    \caption{Standard \text{CAE} based \text{HI} estimates for condition 3 Bearing\_3\_1, Bearing\_3\_2 and Bearing\_3\_3.}
    \label{fig:SAE_HI_B3}
\end{figure*}

Figure~\ref{fig:SAE_HI_B3} illustrates the degradation patterns using the standard \text{CAE} \text{HI} estimation method. Here, \text{HI} estimates might not decrease consistently, even in the training data, highlighting challenges related to the generalizability of this approach. Furthermore, the reconstruction error varies considerably across different bearings, despite operating under similar conditions. This results in varying \text{HI} values, making it difficult to establish a reliable correlation between these values and the actual state of bearing health.

\begin{figure*}[htbp]
    \centering
    \includegraphics[width=15.5cm, height=3.5cm]{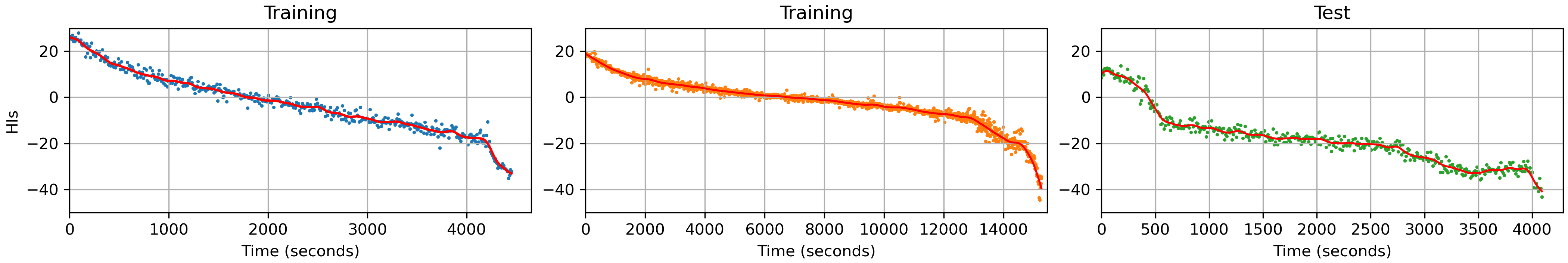}
    \caption{\text{SR-CAE} based \text{HI} estimates for condition 3 Bearing\_3\_1, Bearing\_3\_2 and Bearing\_3\_3.}
    \label{fig:MoAE_HI_B3}
\end{figure*}

Figure~\ref{fig:MoAE_HI_B3} presents the \text{HI} estimates using the \text{SR-CAE} method. This is an alternative approach to enforce monotonicity by regularizing the loss function. In this method, the influences of both the reconstruction and the soft-rank losses on the gradients of the \text{CAE} architecture are equally weighted, with the hyperparameter $\lambda$ set to 1 in Eq.~(\ref{eq:total_loss}). Although this method demonstrates much improved monotonic degradation behavior, there remains considerable variation in \text{HI} values between different bearings. Notably, even in the training sets, substantial differences are observed at both the beginning of the operation and near the failure points.

\begin{figure*}[htbp]
    \centering
    \includegraphics[width=15.5cm, height=3.5cm]{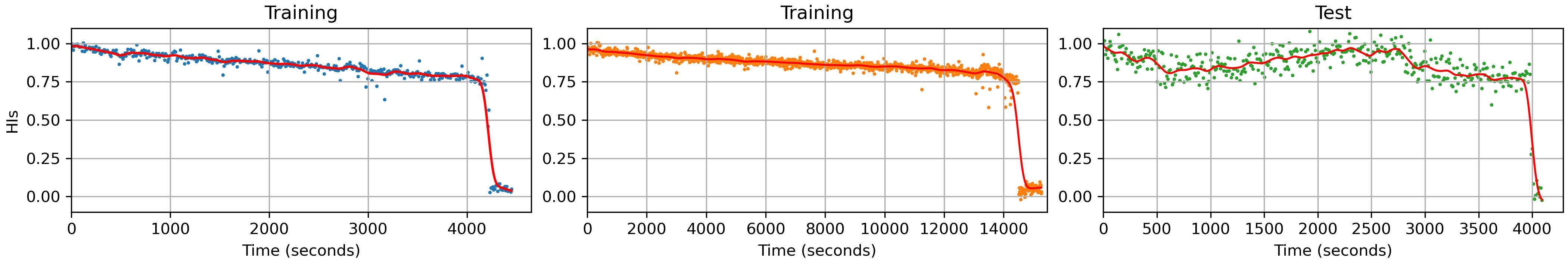}
    \caption{\text{CCAE} based \text{HI} estimates for condition 3 Bearing\_3\_1, Bearing\_3\_2 and Bearing\_3\_3.}
    \label{fig:cG_HI_B3}
\end{figure*}

In the \text{CCAE} method, different rescale factors are applied to the constraints: [1.25, 1.5] for the monotonicity constraint, 1.5 for the energy-\text{HI} consistency constraint and 2.0 for both the upper and lower \text{HI} boundary constraints. Prioritizing boundary constraints ensures that the predicted \text{HI} values remain within the interval [0, 1]. Lowering the priority on boundary constraints increases the chances that the predictions fall outside of this range. The HI estimates based on \text{CCAE} provide several advantages over the baseline methods, as shown in Figure~\ref{fig:cG_HI_B3}. It produces a smoother degradation profile, indicating a more consistent decline in bearing health over time. The \text{HI} values range from 1 (indicating a fully healthy state) to 0 (indicating complete failure), which aligns well with the expected physical degradation process. In addition, the degradation curves correspond to typical bearing degradation, exhibiting a steady decline during the initial phase followed by rapid deterioration as the bearing approaches failure. As a result, the \text{CCAE} approach provides a more accurate and reliable representation of bearing health compared to the baseline methods. 

\subsection{Comparison of Standard \text{CAE} and \text{CCAE} Methods}
\label{subsec:CCAE}
The standard \text{CAE} model is trained using the initial $10\%$ of the training bearing data, which roughly represents the healthy phase of the bearing operation. During this phase, the model reconstruction error is close to zero. However, as the bearing operation time increases, the reconstruction error increases, indicating a degradation in bearing health and a progression towards failure. 

\begin{figure}[htbp]
    \centering
    \includegraphics[width=8.75cm, height=3.0cm]{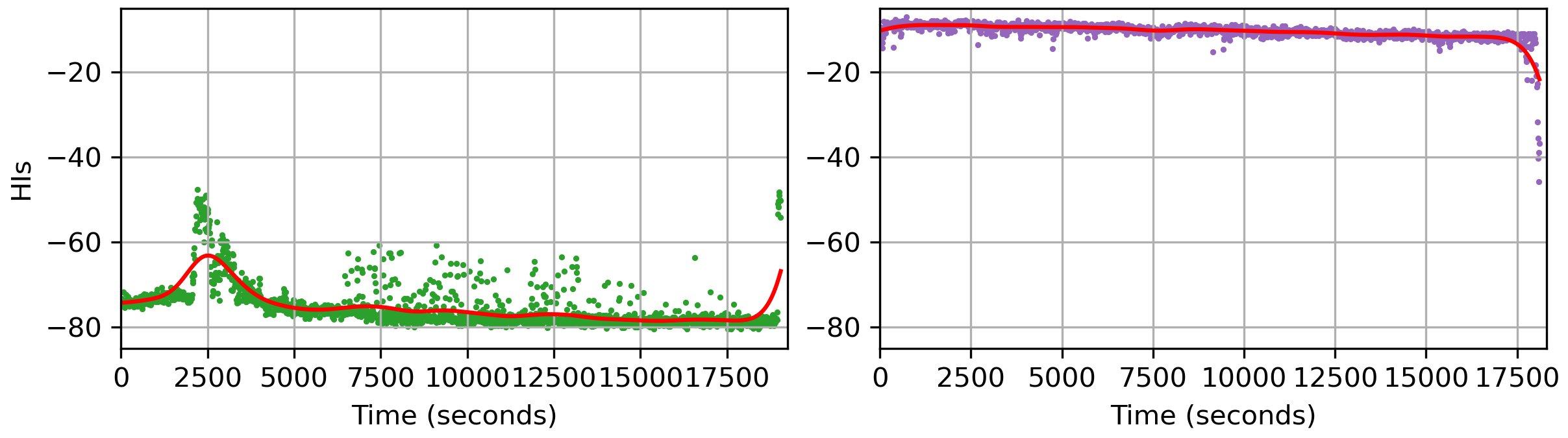}
    \caption{Standard \text{CAE} based \text{HI} estimates for Bearing\_2\_3 and Bearing\_2\_5.}
    \label{fig:CAE_setup2_B3_B5_HIs}
\end{figure}

\begin{figure}[htbp]
    \centering
    \includegraphics[width=8.75cm, height=3.0cm]{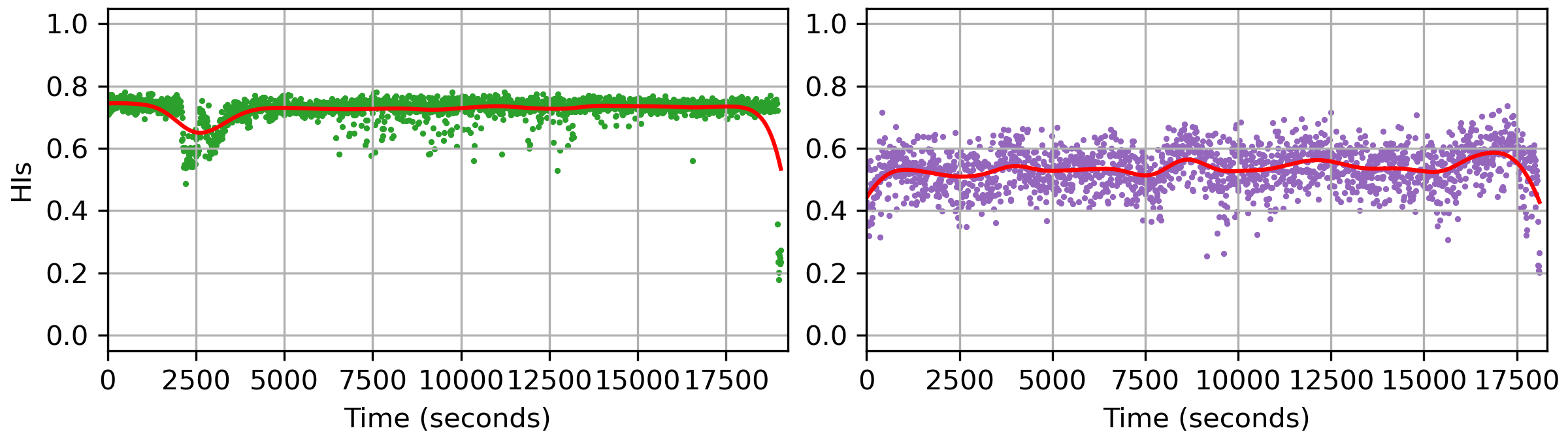}
    \caption{\text{CCAE} based \text{HI} estimates for Bearing\_2\_3 and Bearing\_2\_5.}
    \label{fig:CCAE_setup2_B3_B5_HIs}
\end{figure}

Table~\ref{Tab:Per_evaluation1} summarizes the performance metrics of the standard \text{CAE} method compared to that of the \text{CCAE} method for all the training and test bearings. The results show that \text{CCAE} outperforms the standard \text{CAE} in trendability and consistency by more than $75\%$ and robustness close to $65\%$ considering all the bearings. These results demonstrate that the \text{CCAE} method more effectively tracks the degradation of the bearings over time. It also provides robust, consistent, and repeatable HI estimates across multiple runs. In some cases, while the standard \text{CAE} method might yield superior results, these observations may not fully capture the overall performance of the method. Specifically, for Bearing\_2\_3, the trendability metric value for the standard \text{CAE} method is $-0.732$, which is significantly lower than the $-0.092$ obtained for \text{CCAE}. Although this value suggests a better decreasing correlation of \text{HI} with time, the \text{HI} value actually increases as the bearing approaches failure, contradicting the expected full-run decreasing trend, as depicted in Figure~\ref{fig:CAE_setup2_B3_B5_HIs}. In contrast, the \text{CCAE} method, despite having a lower trendability metric, accurately reflects the expected decrease in \text{HI} values as the bearing approaches failure, as shown in Figure~\ref{fig:CCAE_setup2_B3_B5_HIs}. In addition, for Bearing\_2\_3  the \text{HI} value for the healthiest condition is close to $-50$, approximately 2500 seconds into its operation, and at the end. In contrast, Bearing\_2\_5 approaches an \text{HI} value of $-50$ near failure. This inconsistency, where one bearing's healthiest \text{HI} value is close to another bearing's failure point, underscores the challenge of obtaining a generally representative \text{HI} estimates using the standard \text{CAE} method.

\begin{table*}[htbp]
\renewcommand{\arraystretch}{1.25} 
\begin{center}
\adjustbox{max width=0.9\textwidth}{
\begin{tabular}{ccccccccc}
\hline\hline
\multirow{2}{*}{Bearings} & & \multicolumn{3}{c}{Standard \text{CAE} based \text{HI} Estimation} & & \multicolumn{3}{c}{\text{CCAE} based \text{HI} Estimation} \\ \cline{3-5}\cline{7-9}
  & & Trendability & Robustness & Consistency  & & Trendability & Robustness & Consistency \\ \hline\hline
Bearing\_1\_1 & & $-0.180\pm 0.026$ & $\bm{0.954\pm 0.001}$ & $\bm{0.985\pm 0.012}$ & & $\bm{-0.991\pm 0.005}$ & $0.946\pm 0.003$ & $0.921\pm 0.070$\\ 
Bearing\_1\_2 & & $-0.464\pm 0.055$ & $0.940\pm 0.003$ & $\bm{0.877\pm 0.101}$ & & $\bm{-0.963\pm 0.019}$ & $\bm{0.941\pm 0.005}$ & $0.723\pm 0.244$\\ 
Bearing\_1\_3 & & $0.324\pm 0.027$ & $0.936\pm 0.003$ & $0.691\pm 0.199$ & & $\bm{-0.392\pm 0.394}$ & $\bm{0.946\pm 0.016}$ & $\bm{0.856\pm 0.091}$\\ 
Bearing\_1\_4 & & $-0.772\pm 0.020$ & $0.942\pm 0.001$ & $\bm{0.929\pm 0.056}$ & & $\bm{-0.785\pm 0.109}$ & $\bm{0.948\pm 0.016}$ & $0.814\pm 0.147$\\  
Bearing\_1\_5 & & $0.597\pm 0.077$ & $\bm{0.972\pm 0.002}$ & $0.772\pm 0.171$ & & $\bm{0.316\pm 0.247}$ & $0.946\pm 0.006$ & $\bm{0.922\pm 0.073}$\\ 
Bearing\_1\_6 & & $0.446\pm 0.117$ & $0.916\pm 0.004$ & $0.883\pm 0.089$ & & $\bm{0.157\pm 0.227}$ & $\bm{0.922\pm 0.009}$ & $\bm{0.938\pm 0.059}$\\  
Bearing\_1\_7 & & $\bm{-0.629\pm 0.121}$ & $\bm{0.966\pm 0.002}$ & $0.899\pm 0.083$ & & $0.339\pm 0.248$ & $0.948\pm 0.004$ & $\bm{0.949\pm 0.046}$\\\hline\hline
Bearing\_2\_1 & & $-0.568\pm 0.004$ & $0.927\pm 0.001$ & $0.915\pm 0.025$ & & $\bm{-0.983\pm 0.000}$ & $\bm{0.943\pm 0.004}$ & $\bm{0.922\pm 0.070}$\\ 
Bearing\_2\_2 & & $0.397\pm 0.030$ & $0.948\pm 0.002$ & $0.944\pm 0.025$ & & $\bm{-0.981\pm 0.001}$ & $\bm{0.949\pm 0.006}$ & $\bm{0.959\pm 0.040}$\\ Bearing\_2\_3 & & $\bm{-0.732\pm 0.005}$ & $\bm{0.975\pm 0.008}$ & $0.602\pm 0.224$ & & $-0.092\pm 0.337$ & $0.879\pm 0.091$ & $\bm{0.743\pm 0.184}$\\  
Bearing\_2\_4 & & $0.381\pm 0.108$ & $0.926\pm 0.002$ & $0.816\pm 0.208$ & & $\bm{-0.665\pm 0.143}$ & $\bm{0.930\pm 0.008}$ & $\bm{0.937\pm 0.066}$\\ Bearing\_2\_5 & & $\bm{-0.856\pm 0.051}$ & $\bm{0.921\pm 0.002}$ & $0.850\pm 0.142$ & & $-0.054\pm 0.488$ & $0.874\pm 0.034$ & $\bm{0.896\pm 0.082}$\\  
Bearing\_2\_6 & & $0.683\pm 0.016$ & $0.915\pm 0.005$ & $0.864\pm 0.117$ & & $\bm{-0.220\pm 0.462}$ & $\bm{0.917\pm 0.009}$ & $\bm{0.907\pm 0.083}$\\ 
Bearing\_2\_7 & & $-0.472\pm 0.027$ & $0.846\pm 0.009$ & $0.507\pm 0.346$ & & $\bm{-0.506\pm 0.320}$ & $\bm{0.879\pm 0.107}$ & $\bm{0.600\pm 0.365}$\\\hline\hline   
Bearing\_3\_1 & & $-0.134\pm 0.088$ & $0.930\pm 0.004$ & $0.533\pm 0.343$ & & $\bm{-0.947\pm 0.009}$ & $\bm{0.938\pm 0.005}$ & $\bm{0.946\pm 0.057}$\\ 
Bearing\_3\_2 & & $-0.565\pm 0.078$ & $0.925\pm 0.003$ & $\bm{0.982\pm 0.012}$ & & $\bm{-0.955\pm 0.008}$ & $\bm{0.943\pm 0.004}$ & $0.956\pm 0.045$\\ 
Bearing\_3\_3 & & $\bm{-0.737\pm 0.045}$ & $\bm{0.966\pm 0.001}$ & $0.850\pm 0.102$ & & $-0.599\pm 0.127$ & $0.918\pm 0.013$ & $\bm{0.852\pm 0.143}$\\ \hline\hline
\end{tabular}}
\caption{Standard \text{CAE} Vs. \text{CCAE} based performance evaluations on all bearings.}
\label{Tab:Per_evaluation1}
\end{center}
\end{table*}

\begin{figure}[htbp]
    \centering
    \includegraphics[width=8.75cm, height=3.0cm]{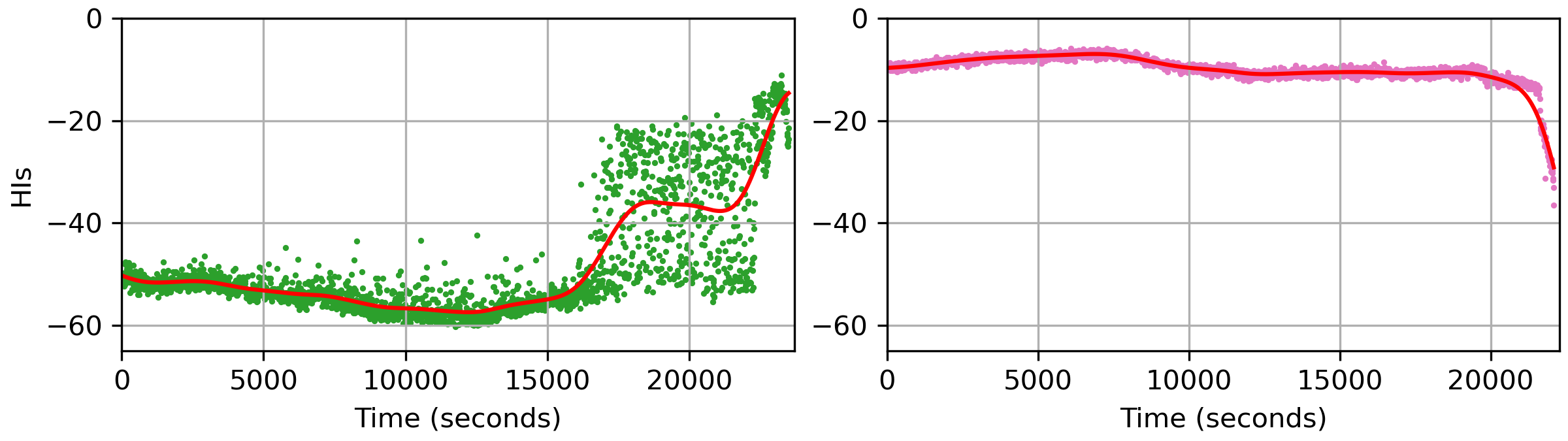}
    \caption{Standard \text{CAE} based \text{HI} estimates for Bearing\_1\_3 and Bearing\_1\_7.}
    \label{fig:CAE_setup1_B3_B7_HIs}
\end{figure}

\begin{figure}[htbp]
    \centering
    \includegraphics[width=8.75cm, height=3.0cm]{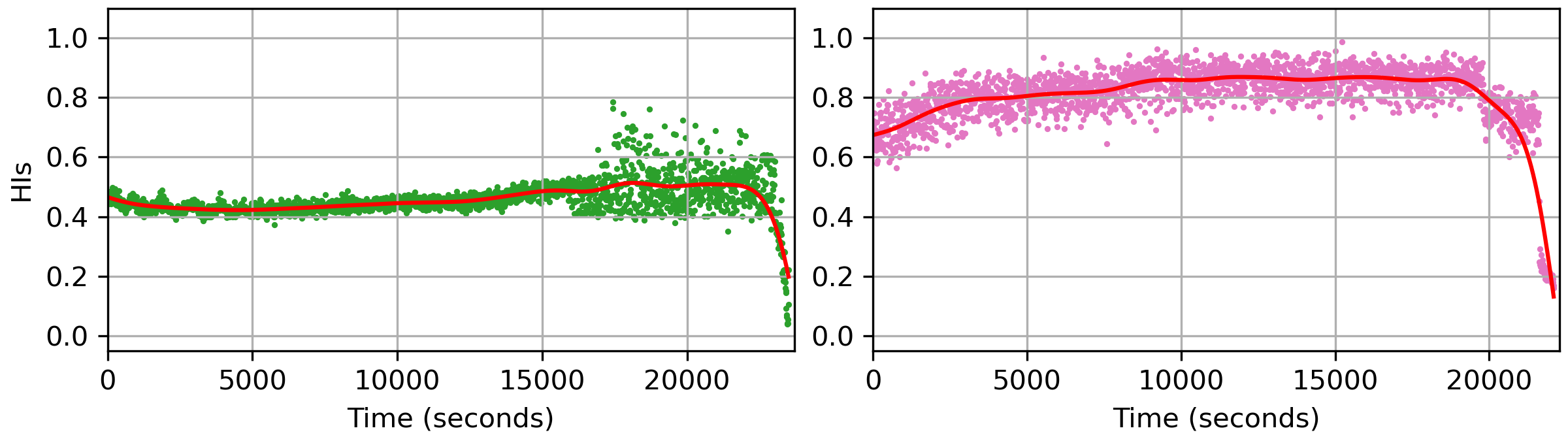}
    \caption{\text{CCAE} based \text{HI} estimates for Bearing\_1\_3 and Bearing\_1\_7.}
    \label{fig:CCAE_setup1_B3_B7_HIs}
\end{figure}

Further insights into performance discrepancies are provided by examining bearings from condition 1 in figures~\ref{fig:CAE_setup1_B3_B7_HIs} and \ref{fig:CCAE_setup1_B3_B7_HIs}. Figure~\ref{fig:Spectral_Energy} shows that Bearing\_1\_3 experiences significant energy fluctuations after 16,000 seconds. These fluctuations are reflected in the \text{HI} estimates of both methods. However, in the case of standard \text{CAE}, the \text{HI} values increase as the failure approach, contrary to expectations of a decreasing trend. However, \text{CCAE} minimizes these fluctuations and maintains \text{HI} values that align with the expected degradation trajectory. For Bearing\_1\_7, the trendability metric for the standard \text{CAE} \text{HI} estimation approach achieves $-0.610$, outperforming the \text{CCAE} score of $0.339$. Despite this higher trendability score for \text{CCAE}, a notable advantage is that its \text{HI} values are bounded within a range of 0 to 1, providing a more interpretable measure of health status.

In general, the \text{CCAE} method improves on the standard \text{CAE} \text{HI} estimation approach by providing more consistent \text{HI} estimates. It is generalizable and robust, handling noisy and unseen test data much better.

\subsection{Comparison of Soft-Rank Loss Function based \text{CAE} and \text{CCAE} Methods}

Table~\ref{Tab:Per_evaluation2} provides a detailed comparison of the performance of the \text{SR-CAE} \text{HI} estimation method compared to the \text{CCAE} method across all bearings. The results indicate that the \text{SR-CAE} method surpasses the \text{CCAE} method in trendability for more than $95\%$ bearings. In contrast, the \text{CCAE} method outperforms the \text{SR-CAE} method in robustness for more than $95\%$ of the bearings and exhibits greater consistency for nearly $90\%$ of the bearings with smaller overall variance.

The advantage of the \text{SR-CAE} method in trendability comes from its objective function, which minimizes the reconstruction loss while enforcing a decreasing monotonicity in the \text{HI} estimates without constraints, resulting in a more monotonic degradation pattern. However, \text{CCAE} demonstrates superior robustness and consistency, showing less variability in \text{HI} estimates. This suggests more stable degradation trends and smoother transitions in \text{HI} values over time. Furthermore, for multiple \text{HI} estimates, the results consistently provide stable and replicable values.

\begin{figure}[htbp]
    \centering
    \includegraphics[width=8.75cm, height=3.0cm]{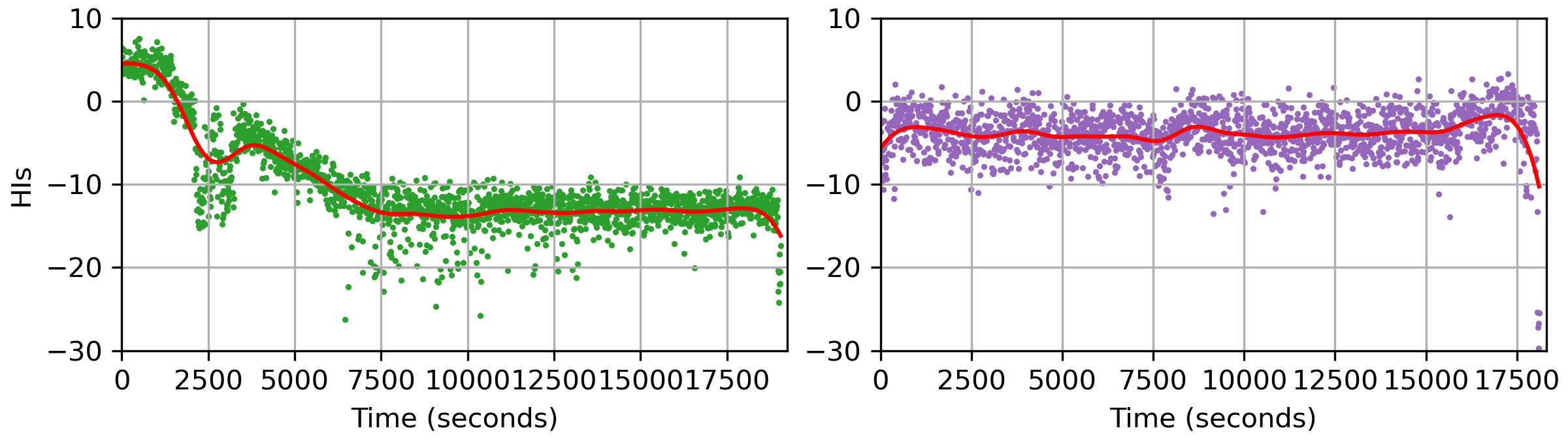}
    \caption{\text{SR-CAE} based \text{HI} estimates for Bearing\_2\_3 and Bearing\_2\_5.}
    \label{fig:SRAE_setup2_B3_B5_HIs}
\end{figure}

Figure~\ref{fig:SRAE_setup2_B3_B5_HIs} provides a validation of these results. The improved monotonicity of the \text{HI} values is notably apparent in Bearing\_2\_3 when using the \text{SR-CAE} method, as opposed to \text{CCAE}. However, significant variability in \text{HI} values is observed in both Bearing\_2\_3 and Bearing\_2\_5 along their lifetime with the \text{SR-CAE} method, unlike \text{CCAE} shown in Figure~\ref{fig:CCAE_setup2_B3_B5_HIs}. Such disparities complicate the generalizability of the \text{SR-CAE} method when compared to the \text{CCAE} method.

\begin{table*}[htbp]
\renewcommand{\arraystretch}{1.25} 
\begin{center}
\adjustbox{max width=0.9\textwidth}{
\begin{tabular}{ccccccccc}
\hline\hline
\multirow{2}{*}{Bearings} & & \multicolumn{3}{c}{\text{SR-CAE} based \text{HI} Estimation} & & \multicolumn{3}{c}{\text{CCAE} based \text{HI} Estimation}\\ \cline{3-5}\cline{7-9}
  & & Trendability & Robustness & Consistency  & & Trendability & Robustness & Consistency \\ \hline\hline
Bearing\_1\_1 & & $\bm{-0.998\pm 0.000}$ & $0.895\pm 0.009$ & $0.915\pm 0.047$ & & $-0.991\pm 0.005$ & $\bm{0.946\pm 0.003}$ & $\bm{0.921\pm 0.070}$\\ 
Bearing\_1\_2 & & $\bm{-0.989\pm 0.001}$ & $0.859\pm 0.021$ & $\bm{0.860\pm 0.076}$ & & $-0.963\pm 0.019$ & $\bm{0.941\pm 0.005}$ & $0.723\pm 0.244$\\ 
Bearing\_1\_3 & & $\bm{-0.418\pm 0.147}$ & $0.906\pm 0.011$ & $0.762\pm 0.113$ & & $-0.392\pm 0.394$ & $\bm{0.946\pm 0.016}$ & $\bm{0.856\pm 0.091}$\\ 
Bearing\_1\_4 & & $\bm{-0.921\pm 0.025}$ & $0.867\pm 0.047$ & $\bm{0.844\pm 0.149}$ & & $-0.785\pm 0.109$ & $\bm{0.948\pm 0.016}$ & $0.814\pm 0.147$\\  
Bearing\_1\_5 & & $\bm{-0.537\pm 0.174}$ & $0.552\pm 0.099$ & $0.913\pm 0.067$ & & $0.316\pm 0.247$ & $\bm{0.946\pm 0.006}$ & $\bm{0.922\pm 0.073}$\\ 
Bearing\_1\_6 & & $\bm{-0.500\pm 0.130}$ & $0.812\pm 0.054$ & $0.901\pm 0.114$ & & $0.157\pm 0.227$ & $\bm{0.922\pm 0.009}$ & $\bm{0.938\pm 0.059}$\\ 
Bearing\_1\_7 & & $\bm{-0.868\pm 0.021}$ & $0.636\pm 0.026$ & $0.930\pm 0.055$ & & $0.339\pm 0.248$ & $\bm{0.948\pm 0.004}$ & $\bm{0.949\pm 0.046}$\\\hline\hline
Bearing\_2\_1 & & $\bm{-0.995\pm 0.001}$ & $0.896\pm 0.008$ & $0.825\pm 0.113$ & & $-0.983\pm 0.000$ & $\bm{0.943\pm 0.004}$ & $\bm{0.922\pm 0.070}$\\ 
Bearing\_2\_2 & & $\bm{-0.997\pm 0.000}$ & $0.893\pm 0.005$ & $0.917\pm 0.057$ & & $-0.981\pm 0.001$ & $\bm{0.949\pm 0.006}$ & $\bm{0.959\pm 0.040}$\\  
Bearing\_2\_3 & & $\bm{-0.672\pm 0.054}$ & $\bm{0.938\pm 0.034}$ & $0.666 \pm 0.201$ & & $-0.092\pm 0.337$ & $0.879\pm 0.091$ & $\bm{0.743\pm 0.184}$\\ 
Bearing\_2\_4 & & $\bm{-0.839\pm 0.040}$ & $0.627\pm 0.048$ & $0.789\pm 0.167$ & & $-0.665\pm 0.143$ & $\bm{0.930\pm 0.008}$ & $\bm{0.937\pm 0.066}$\\  
Bearing\_2\_5 & & $0.009\pm 0.156$ & $0.710\pm 0.086$ & $0.878\pm 0.128$ & & $\bm{-0.054\pm 0.488}$ & $\bm{0.874\pm 0.037}$ & $\bm{0.896\pm 0.082}$\\ 
Bearing\_2\_6 & & $\bm{-0.500\pm 0.201}$ & $0.774\pm 0.046$ & $0.829\pm 0.161$ & & $-0.220\pm 0.462$ & $\bm{0.917\pm 0.009}$ & $\bm{0.907\pm 0.083}$\\  
Bearing\_2\_7 & & $\bm{-0.803\pm 0.071}$ & $0.845\pm 0.112$ & $0.423\pm 0.332$ & & $-0.506\pm 0.320$ & $\bm{0.879\pm 0.107}$ & $\bm{0.600\pm 0.365}$\\\hline\hline 
Bearing\_3\_1 & & $\bm{-0.988\pm 0.002}$ & $0.805\pm 0.004$ & $0.927\pm 0.032$ &  & $-0.947\pm 0.009$ & $\bm{0.938\pm 0.005}$ & $\bm{0.946\pm 0.057}$\\ 
Bearing\_3\_2 & & $\bm{-0.995\pm 0.001}$ & $0.838\pm 0.012$ & $0.932\pm 0.056$ & & $-0.955\pm 0.008$ & $\bm{0.943\pm 0.004}$ & $\bm{0.956\pm 0.045}$\\ 
Bearing\_3\_3 & & $\bm{-0.966\pm 0.005}$ & $0.884\pm 0.017$ & $0.805\pm 0.135$ &  & $-0.599\pm 0.127$ & $\bm{0.918\pm 0.013}$ & $\bm{0.852\pm 0.143}$\\ \hline\hline
\end{tabular}}
\caption{\text{SR-CAE} Vs. \text{CCAE} performance evaluations on all bearings.}
\label{Tab:Per_evaluation2}
\end{center}
\end{table*}

\section{Ablation study}
\label{sec:ablation}
In this section, we analyze the effects of modifying various aspects of the complete implementation of the \text{CCAE} method. Specifically, we examine the impact of individual constraints within the \text{CCAE} method, the influence of different constraint rescale factors, and the effect of replacing the monotonicity constraint with a soft-rank loss function in the \text{CCAE} method.

\subsection{Impact of Constraints on \text{CCAE}}
In this section, we examine the impact of each constraint in the \text{CCAE} implementation by systematically excluding one constraint at a time and evaluating the resulting performance. The experiments devised for this study are the following:
\begin{enumerate}
    \item \textbf{\text{CCAE\_EB}}: The monotonicity constraint is excluded. However, the energy-\text{HI} consistency and boundary constraints are retained, with rescale factors of 1.5 and 2.0, respectively.
    \item \textbf{\text{CCAE\_MB}}: The energy-\text{HI} consistency constraint is excluded. However, monotonicity and boundary constraints are retained, with rescale factors of [1.25, 1.5] and 2.0, respectively.
    \item \textbf{\text{CCAE\_ME}}: The boundary constraints are excluded, leaving only the monotonicity and energy-\text{HI} consistency constraints, with rescale factors of [1.25, 1.5] and 1.5, respectively.
\end{enumerate}

The results of this experiment, presented in Table~\ref{Tab:Constraints_Impact}, highlight the importance of the different constraints in the \text{CCAE} method. The monotonicity constraint significantly enhances the trendability of the \text{HI} estimates, as its primary objective is to enforce the progressive health degradation of the bearings. The boundary constraint plays a crucial role in maintaining the consistency of the \text{HI} estimates by ensuring that they remain within the defined range of [0, 1], leading to reliable \text{HI} estimates across multiple experiments. Furthermore, the energy-\text{HI} consistency constraint contributes to both robustness and consistency. Its formulation penalizes high fluctuations in \text{HI} predictions that do not align with the energy progression of the signal, thereby providing a smooth and reliable degradation trend over time. Ultimately, all constraints play a vital role in enforcing the desired properties expected of the bearing \text{HI} estimates. By incorporating all these constraints, the \text{CCAE} approach effectively captures the degradation pattern while ensuring robustness and consistency in the \text{HI} estimates.

\begin{table*}[htbp]
\renewcommand{\arraystretch}{1.5} 
\begin{center}
\adjustbox{max width=1.0\textwidth}{
\begin{tabular}{cccccccccccccccc}
\hline\hline
\multirow{2}{*}{Bearings} & & \multicolumn{3}{c}{\text{CCAE\_EB}} & & \multicolumn{3}{c}{\text{CCAE\_MB}} & &\multicolumn{3}{c}{\text{CCAE\_ME}}\\ \cline{3-5}\cline{7-9}\cline{11-13}
& & Trendability & Robustness & Consistency & & Trendability & Robustness & Consistency & & Trendability & Robustness & Consistency\\ \hline\hline
Bearing\_1\_1 & & $-0.523\pm 0.075$ & $0.923\pm 0.005$ & $\bm{0.995\pm 0.004}$ & & $-0.985\pm 0.003$ & $0.904\pm 0.012$ & $0.958\pm 0.036$ & & $\bm{-0.994\pm 0.002}$ & $\bm{0.991\pm 0.001}$ & $0.862\pm 0.087$\\
Bearing\_1\_2 & & $-0.557\pm 0.054$ & $0.930\pm 0.009$ & $0.877\pm 0.147$ & & $-0.876\pm 0.027$ & $0.915\pm 0.006$ & $\bm{0.935\pm 0.060}$ & & $\bm{-0.966\pm 0.011}$ & $\bm{0.988\pm 0.002}$ & $0.495\pm 0.293$\\
Bearing\_1\_3 & & $-0.188\pm 0.320$ & $0.875\pm 0.131$ & $0.814\pm 0.112$ & & $\bm{-0.775\pm 0.205}$ & $0.923\pm 0.020$ & $\bm{0.911\pm 0.052}$ & & $-0.064\pm 0.300$ & $\bm{0.956\pm 0.011}$ & $0.909\pm 0.069$\\
Bearing\_1\_4 & & $-0.704\pm 0.133$ & $0.909\pm 0.027$ & $0.855\pm 0.107$ & & $\bm{-0.710\pm 0.224}$ & $0.950\pm 0.010$ & $\bm{0.895\pm 0.072}$ & & $-0.629\pm 0.177$ & $\bm{0.972\pm 0.005}$ & $0.862\pm 0.090$\\
Bearing\_1\_5 & & $0.230\pm 0.240$ & $0.918\pm 0.020$ & $0.864\pm 0.187$ & & $\bm{-0.237\pm 0.228}$ & $0.934\pm 0.011$ & $\bm{0.864\pm 0.142}$ & & $-0.046\pm 0.296$ & $\bm{0.974\pm 0.005}$ & $0.838\pm 0.140$\\
Bearing\_1\_6 & & $0.034\pm 0.153$ & $0.892\pm 0.017$ & $\bm{0.927\pm 0.082}$ & & $\bm{-0.433\pm 0.142}$ & $0.896\pm 0.031$ & $0.860\pm 0.167$ & & $0.028\pm 0.123$ & $\bm{0.968\pm 0.007}$ & $0.833\pm 0.181$\\
Bearing\_1\_7 & & $0.407\pm 0.172$ & $0.923\pm 0.010$ & $\bm{0.939\pm 0.062}$ & & $\bm{-0.658\pm 0.204}$ & $0.935\pm 0.011$ & $0.907\pm 0.075$ & & $-0.143\pm 0.388$ & $\bm{0.973\pm 0.004}$ & $0.929\pm 0.055$\\\hline\hline
Bearing\_2\_1 & & $-0.622\pm 0.060$ & $0.897\pm 0.006$ & $\bm{0.993\pm 0.005}$ & & $\bm{-0.991\pm 0.002}$ & $0.937\pm 0.005$ & $0.868\pm 0.128$ & & $-0.983\pm 0.003$ & $\bm{0.975\pm 0.004}$ & $0.671\pm 0.224$ \\ 
Bearing\_2\_2 & & $-0.598\pm 0.045$ & $0.900\pm 0.006$ & $\bm{0.992\pm 0.005}$ & & $\bm{-0.995\pm 0.001}$ & $0.944\pm 0.009$ & $0.901\pm 0.093$ & & $-0.987\pm 0.005$ & $\bm{0.977\pm 0.006}$ & $0.863\pm 0.123$ \\ 
Bearing\_2\_3 & & $-0.264\pm 0.308$ & $0.817\pm 0.169$ & $\bm{0.747\pm 0.168}$ & & $\bm{-0.402\pm 0.192}$ & $\bm{0.858\pm 0.097}$ & $0.652\pm 0.197$ & & $-0.323\pm 0.201$ & $0.853\pm 0.100$ & $0.682\pm 0.191$ \\  
Bearing\_2\_4 & & $-0.638\pm 0.377$ & $0.905\pm 0.027$ & $\bm{0.893\pm 0.076}$ & & $-0.715\pm 0.186$ & $0.925\pm 0.011$ & $0.746\pm 0.254$ & & $\bm{-0.765\pm 0.120}$ & $\bm{0.942\pm 0.012}$ & $0.732\pm 0.230$ \\   
Bearing\_2\_5 & & $0.166\pm 0.306$ & $0.808\pm 0.080$ & $\bm{0.918\pm 0.084}$ & & $\bm{-0.180\pm 0.283}$ & $0.869\pm 0.044$ & $0.864\pm 0.127$ & & $-0.130\pm 0.318$ & $\bm{0.916\pm 0.045}$ & $0.856\pm 0.139$ \\   
Bearing\_2\_6 & & $-0.168\pm 0.322$ & $0.883\pm 0.025$ & $0.899\pm 0.104$ & & $-0.387\pm 0.287$ & $0.905\pm 0.014$ & $0.847\pm 0.139$ & & $\bm{-0.571\pm 0.211}$ & $\bm{0.937\pm 0.008}$ & $\bm{0.923\pm 0.081}$ \\   
Bearing\_2\_7 & & $-0.486\pm 0.234$ & $0.882\pm 0.080$ & $\bm{0.563\pm 0.287}$ & & $\bm{-0.611\pm 0.262}$ & $0.841\pm 0.127$ & $0.393\pm 0.410$ & & $-0.518\pm 0.277$ & $\bm{0.892\pm 0.088}$ & $0.474\pm 0.353$ \\ \hline\hline
Bearing\_3\_1 & & $-0.484\pm 0.038$ & $0.915\pm 0.011$ & $\bm{0.969\pm 0.033}$ & & $\bm{-0.981\pm 0.003}$ & $0.932\pm 0.006$ & $0.945\pm 0.046$ & & $-0.959\pm 0.008$ & $\bm{0.983\pm 0.003}$ & $0.904\pm 0.104$\\ 
Bearing\_3\_2 & & $-0.496\pm 0.027$ & $0.917\pm 0.003$ & $\bm{0.992\pm 0.006}$ & & $\bm{-0.991\pm 0.002}$ & $0.951\pm 0.006$ & $0.923\pm 0.063$ & & $-0.974\pm 0.003$ & $\bm{0.986\pm 0.004}$ & $0.915\pm 0.062$\\ 
Bearing\_3\_3 & & $-0.234\pm 0.317$ & $0.910\pm 0.016$ & $0.898\pm 0.101$ & & $\bm{-0.769\pm 0.151}$ & $0.855\pm 0.030$ & $\bm{0.949\pm 0.040}$ & & $-0.673\pm 0.231$ & $\bm{0.954\pm 0.012}$ & $0.928\pm 0.069$\\\hline\hline
\end{tabular}}
\caption{Impact of constraints on \text{CCAE} performance on all bearings.}
\label{Tab:Constraints_Impact}
\end{center}
\end{table*}

\subsection{Effects of Rescale Factors on \text{CCAE}}
This section examines the impact of the selection of the rescale factors on the performance of \text{CCAE}. The objective of this experiment is to assess the stability of the \text{CCAE} method, demonstrating that small changes to the rescale factors do not lead to significant performance fluctuations. Building on the results from Section~\ref{subsec:CCAE}, where the rescale factors [1.25, 1.5], 1.5, 2.0, and 2.0 (referred as \text{RF\_C1}) were applied to the monotonicity, energy-\text{HI} consistency, upper and lower boundary constraints, we conduct additional experiment to evaluate the effect of small variations in these factors. To this end, we examine an alternative set of rescale factors: [1.05, 1.25], 1.25, 1.25, and 1.25 (referred as \text{RF\_C2}).

\begin{table*}[htbp]
\renewcommand{\arraystretch}{1.25} 
\begin{center}
\adjustbox{max width=0.9\textwidth}{
\begin{tabular}{ccccccccc}
\hline\hline
\multirow{2}{*}{Bearings} & & \multicolumn{3}{c}{\text{RF\_C1}} & & \multicolumn{3}{c}{\text{RF\_C2}} \\ \cline{3-5}\cline{7-9}
& & Trendability & Robustness & Consistency  & & Trendability & Robustness & Consistency \\ \hline\hline
Bearing\_1\_1 & & $\bm{-0.993\pm 0.002}$ & $0.944\pm 0.004$ & $0.908\pm 0.071$ & & $-0.991\pm 0.005$ & $\bm{0.946\pm 0.003}$ & $\bm{0.921\pm 0.070}$\\ 
Bearing\_1\_2 & & $\bm{-0.971\pm 0.008}$ & $0.940\pm 0.004$ & $\bm{0.804\pm 0.111}$ & & $-0.963\pm 0.019$ & $\bm{0.941\pm 0.005}$ & $0.723\pm 0.244$\\ 
Bearing\_1\_3 & & $-0.313\pm 0.408$ & $\bm{0.954\pm 0.006}$ & $\bm{0.864\pm 0.118}$ & & $\bm{-0.392\pm 0.394}$ & $0.946\pm 0.016$ & $0.856\pm 0.091$\\ 
Bearing\_1\_4 & & $-0.741\pm 0.188$ & $\bm{0.952\pm 0.007}$ & $\bm{0.878\pm 0.093}$ & & $\bm{-0.785\pm 0.109}$ & $0.948\pm 0.016$ & $0.814\pm 0.147$\\  
Bearing\_1\_5 & & $0.331\pm 0.146$ & $\bm{0.948\pm 0.006}$ & $\bm{0.946\pm 0.056}$ & & $\bm{0.316\pm 0.247}$ & $0.946\pm 0.006$ & $0.922\pm 0.073$\\ 
Bearing\_1\_6 & & $\bm{0.108\pm 0.146}$ & $\bm{0.925\pm 0.017}$ & $0.839\pm 0.119$ & & $0.157\pm 0.227$ & $0.922\pm 0.009$ & $\bm{0.938\pm 0.059}$\\  
Bearing\_1\_7 & & $0.359\pm 0.148$ & $\bm{0.950\pm 0.005}$ & $0.920\pm 0.073$ & & $\bm{0.339\pm 0.248}$ & $0.948\pm 0.004$ & $\bm{0.949\pm 0.046}$\\\hline\hline  
Bearing\_2\_1 & & $\bm{-0.984\pm 0.002}$ & $0.939\pm 0.005$ & $\bm{0.936\pm 0.078}$ & & $-0.983\pm 0.000$ & $\bm{0.943\pm 0.004}$ & $0.922\pm 0.070$\\ 
Bearing\_2\_2 & & $-0.981\pm 0.005$ & $0.942\pm 0.003$ & $0.950\pm 0.038$ & & $\bm{-0.981\pm 0.001}$ & $\bm{0.949\pm 0.006}$ & $\bm{0.959\pm 0.040}$\\   
Bearing\_2\_3 & & $\bm{-0.322\pm 0.228}$ & $\bm{0.917\pm 0.017}$ & $0.737\pm 0.177$ & & $-0.092\pm 0.337$ & $0.879\pm 0.091$ & $\bm{0.743\pm 0.184}$\\   
Bearing\_2\_4 & & $-0.640\pm 0.211$ & $0.924\pm 0.009$ & $0.850\pm 0.168$ & & $\bm{-0.665\pm 0.143}$ & $\bm{0.930\pm 0.008}$ & $\bm{0.937\pm 0.066}$\\  
Bearing\_2\_5 & & $\bm{-0.165\pm 0.351}$ & $0.871\pm 0.030$ & $0.886\pm 0.100$ & & $-0.054\pm 0.488$ & $\bm{0.874\pm 0.037}$ & $\bm{0.896\pm 0.082}$\\  
Bearing\_2\_6 & & $-0.035\pm 0.291$ & $\bm{0.919\pm 0.006}$ & $0.879\pm 0.147$ & & $\bm{-0.220\pm 0.462}$ & $0.917\pm 0.009$ & $\bm{0.907\pm 0.083}$\\   
Bearing\_2\_7 & & $-0.474\pm 0.267$ & $\bm{0.922\pm 0.018}$ & $0.553\pm 0.324$ & & $\bm{-0.506\pm 0.320}$ & $0.879\pm 0.107$ & $\bm{0.600\pm 0.365}$\\\hline\hline  
Bearing\_3\_1 & & $\bm{-0.948\pm 0.010}$ & $\bm{0.941\pm 0.009}$ & $0.936\pm 0.063$ &  & $-0.947\pm 0.009$ & $0.938\pm 0.005$ & $\bm{0.946\pm 0.057}$\\ 
Bearing\_3\_2 & & $-0.948\pm 0.013$ & $0.942\pm 0.005$ & $0.938\pm 0.051$ & & $\bm{-0.955\pm 0.008}$ & $\bm{0.943\pm 0.004}$ & $\bm{0.956\pm 0.045}$\\ 
Bearing\_3\_3 & & $-0.554\pm 0.229$ & $\bm{0.922\pm 0.009}$ & $\bm{0.925\pm 0.063}$ & & $\bm{-0.599\pm 0.127}$ & $0.918\pm 0.013$ & $0.852\pm 0.143$\\\hline\hline
\end{tabular}}
\caption{Effects of rescale factors on \text{CCAE} performance on all bearings.}
\label{Tab:RF_Impact}
\end{center}
\end{table*}

The results of this experiment, presented in Table~\ref{Tab:RF_Impact}, highlight the performance differences of the \text{CCAE} method when using different constraint rescale factors. The findings indicate that in terms of trendability, \text{RF\_C2} slightly outperforms \text{RF\_C1} in $58.82\%$ of the bearings. Both models show minimal variability, with the largest observed differences being $-0.322\pm 0.228$ for \text{RF\_C1} and $-0.092\pm 0.337$ for \text{RF\_C2} in Bearing\_2\_3. When comparing robustness, \text{RF\_C1} outperforms \text{RF\_C2} in $58.82\%$ of the bearings. This suggests a slight overall advantage for \text{RF\_C1}. Furthermore, the variability between the models for robustness remains minimal, with the largest differences being $0.922 \pm 0.018$ for \text{RF\_C1} and $0.879 \pm 0.107$ for \text{RF\_C2} in Bearing\_2\_7.  In terms of consistency, \text{RF\_C2} demonstrates superior performance over \text{RF\_C1} in $64.71\%$ of the bearings. This improvement is attributed to the larger boundary constraint rescale factor used in \text{RF\_C2}. The largest differences in consistency are observed in Bearing\_1\_6, where \text{RF\_C1} achieves $0.839\pm 0.119$ and \text{RF\_C2} has a value of $0.938\pm 0.059$.

Overall, the results indicate that the performance of both approaches is comparable, and small changes in rescale factors do not lead to significant performance fluctuations.

\subsection{Soft Rank based \text{CCAE}}
In this approach, the modification of \text{CCAE} involves replacing the monotonic degradation constraint with a soft-ranking in the loss function. The influences of the reconstruction and the soft-rank losses on the gradients of the model architecture are equally weighted, with a value of $\lambda$ set to 1 in Eq.~(\ref{eq:total_loss}). Additionally, only the upper and lower bounds are included as constraints, with a rescale factor set to 2.0 and excluding the energy-\text{HI} consistency constraint.

Table~\ref{Tab:SoftRank_CCAE} summarizes the performance comparison between the soft-rank \text{CCAE} method and the \text{CCAE} method with all constraints across all bearings. The results show that the soft-rank \text{CCAE} method outperforms the \text{CCAE} method in $82.36\%$ of the bearings based on the trendability metric. This superior performance is especially clear across all test bearings. Although both methods integrate monotonicity behavior differently, the soft-rank \text{CCAE} method benefits from having one fewer constraint to satisfy the energy-HI consistency requirement, giving it a clear advantage in the trendability metric. Furthermore, the \text{CCAE} method exhibits a $100\%$ advantage in terms of robustness across all bearings, mainly due to the inclusion of the energy-\text{HI} consistency constraint. The \text{CCAE} method exhibits a higher level of consistency than the soft-rank \text{CCAE} method, showing a $64.71\%$ improvement across all bearings and performing particularly well on the test bearings. In summary, explicitly adding desired properties as constraints during training is advantageous for enhancing model performance. Although adding a particular constraint may decrease a single metric, it can lead to overall improvement when considering all metrics collectively.

\begin{table*}[htbp]
\renewcommand{\arraystretch}{1.25} 
\begin{center}
\adjustbox{max width=0.9\textwidth}{
\begin{tabular}{ccccccccc}
\hline\hline
\multirow{2}{*}{Bearings} & & \multicolumn{3}{c}{Soft Rank \text{CCAE} based \text{HI} Estimation} & & \multicolumn{3}{c}{\text{CCAE} based \text{HI} Estimation} \\ \cline{3-5}\cline{7-9}
& & Trendability & Robustness & Consistency  & & Trendability & Robustness & Consistency \\ \hline\hline
Bearing\_1\_1 & & $-0.972\pm 0.009$ & $0.754\pm 0.025$ & $\bm{0.986\pm 0.008}$ & & $\bm{-0.991\pm 0.005}$ & $\bm{0.946\pm 0.003}$ & $0.921\pm 0.070$\\ 
Bearing\_1\_2 & & $\bm{-0.970\pm 0.003}$ & $0.744\pm 0.014$ & $\bm{0.932\pm 0.066}$ & & $-0.963\pm 0.019$ & $\bm{0.941\pm 0.005}$ & $0.723\pm 0.244$\\ 
Bearing\_1\_3 & & $\bm{-0.893\pm 0.066}$ & $0.769\pm 0.080$ & $0.828\pm 0.121$ & & $-0.392\pm 0.394$ & $\bm{0.946\pm 0.016}$ & $\bm{0.856\pm 0.091}$\\ 
Bearing\_1\_4 & & $\bm{-0.905\pm 0.035}$ & $0.869\pm 0.032$ & $\bm{0.840\pm 0.118}$ & & $-0.785\pm 0.109$ & $\bm{0.948\pm 0.016}$ & $0.814\pm 0.147$\\  
Bearing\_1\_5 & & $\bm{-0.093\pm 0.254}$ & $0.816\pm 0.061$ & $0.882\pm 0.115$ & & $0.316\pm 0.247$ & $\bm{0.946\pm 0.006}$ & $\bm{0.922\pm 0.073}$\\ 
Bearing\_1\_6 & & $\bm{-0.077\pm 0.192}$ & $0.775\pm 0.126$ & $0.778\pm 0.210$ & & $0.157\pm 0.227$ & $\bm{0.922\pm 0.009}$ & $\bm{0.938\pm 0.059}$\\  
Bearing\_1\_7 & & $\bm{-0.552\pm 0.146}$ & $0.883\pm 0.029$ & $0.843\pm 0.138$ & & $0.339\pm 0.248$ & $\bm{0.948\pm 0.004}$ & $\bm{0.949\pm 0.046}$\\\hline\hline 
Bearing\_2\_1 & & $-0.963\pm 0.005$ & $0.700\pm 0.007$ & $\bm{0.976\pm 0.024}$ & & $\bm{-0.983\pm 0.000}$ & $\bm{0.943\pm 0.004}$ & $0.922\pm 0.070$\\ 
Bearing\_2\_2 & & $-0.970\pm 0.005$ & $0.721\pm 0.016$ & $\bm{0.978\pm 0.020}$ & & $\bm{-0.981\pm 0.001}$ & $\bm{0.949\pm 0.006}$ & $0.959\pm 0.040$\\   
Bearing\_2\_3 & & $\bm{-0.403\pm 0.209}$ & $0.749\pm 0.195$ & $0.679\pm 0.200$ & & $-0.092\pm 0.337$ & $\bm{0.879\pm 0.091}$ & $\bm{0.743\pm 0.184}$\\   
Bearing\_2\_4 & & $\bm{-0.874\pm 0.051}$ & $0.801\pm 0.063$ & $0.751\pm 0.266$ & & $-0.665\pm 0.143$ & $\bm{0.930\pm 0.008}$ & $\bm{0.937\pm 0.066}$\\  
Bearing\_2\_5 & & $\bm{-0.273\pm 0.330}$ & $0.637\pm 0.143$ & $0.819\pm 0.132$ & & $-0.054\pm 0.488$ & $\bm{0.874\pm 0.037}$ & $\bm{0.896\pm 0.082}$\\  
Bearing\_2\_6 & & $\bm{-0.408\pm 0.251}$ & $0.785\pm 0.083$ & $0.885\pm 0.146$ & & $-0.220\pm 0.462$ & $\bm{0.917\pm 0.009}$ & $\bm{0.907\pm 0.083}$\\   
Bearing\_2\_7 & & $\bm{-0.765\pm 0.083}$ & $0.776\pm 0.105$ & $0.310\pm 0.353$ & & $-0.506\pm 0.320$ & $\bm{0.879\pm 0.107}$ & $\bm{0.600\pm 0.365}$\\\hline\hline   
Bearing\_3\_1 & & $\bm{-0.967\pm 0.004}$ & $0.751\pm 0.016$ & $0.937\pm 0.054$ & & $-0.947\pm 0.009$ & $\bm{0.938\pm 0.005}$ & $\bm{0.946\pm 0.057}$\\ 
Bearing\_3\_2 & & $\bm{-0.964\pm 0.007}$ & $0.688\pm 0.023$ & $\bm{0.984\pm 0.013}$ & & $-0.955\pm 0.008$ & $\bm{0.943\pm 0.004}$ & $0.956\pm 0.045$\\ 
Bearing\_3\_3 & & $\bm{-0.908\pm 0.028}$ & $0.690\pm 0.052$ & $0.837\pm 0.118$ & & $-0.599\pm 0.127$ & $\bm{0.918\pm 0.013}$ & $\bm{0.852\pm 0.143}$\\\hline\hline
\end{tabular}}
\caption{Soft Rank \text{CCAE} Vs. \text{CCAE} based performance evaluations on all bearings.}
\label{Tab:SoftRank_CCAE}
\end{center}
\end{table*} 

\section{Conclusion}
\label{sec:conclusion}
In conclusion, this study successfully demonstrates the potential of a constraint-guided deep learning framework, specifically \text{CCAE}, to develop physically consistent health indicators for bearing \text{PHM}. By incorporating domain knowledge through monotonicity, boundary and energy-\text{HI} consistency constraints, the \text{CCAE} addresses the limitations of both conventional data-driven and physics-based methods. The experimental results show that \text{CCAE} is significantly better than the standard \text{CAE}, achieving $65\%$ higher robustness and $75\%$ higher consistency. It also outperforms the \text{SR-CAE} baseline with a $95\%$ improvement in robustness and a $90\%$ improvement in consistency. Although the \text{SR-CAE} method excels in trendability metrics for $95\%$ of bearings due to the explicit use of monotonicity in model training, \text{CCAE} performs better than \text{CAE} for $75\%$ of the bearings. Moreover, the \text{CCAE} generates \text{HI} outputs that are bounded within a specified range and reliably represent the bearing's health state. Its degradation profiles are smoother and closely align with expected physical degradation patterns, underscoring the value of incorporating domain knowledge constraints.
The ablation study further confirms that the monotonicity constraint enhances trendability, the boundary constraint ensures consistency, and the energy-\text{HI} consistency constraint improves robustness. Furthermore, our findings indicate that minor changes in the rescale factor of these constraints do not substantially affect the performance of \text{CCAE}. These findings underscore the potential of employing representative constraints in the proposed deep learning framework to generate reliable bearing \text{HI}s. Future research could explore the performance of the framework in other datasets, incorporate additional domain-specific constraints, investigate its application to other areas with tailored domain constraints, and examine its potential for \text{RUL} prediction. This approach could also be extended to handle more complex degradation scenarios, building on the methodology presented here. Overall, this research offers a promising direction for future prognostic applications, improving the reliability and effectiveness of asset health management strategies.


\bibliographystyle{apacite}
\PHMbibliography{ijphm}

\end{document}